\documentclass[manuscript]{acmart}
    
\usepackage{amsmath,amsfonts}
\usepackage{algorithmic}
\usepackage{algorithm}
\usepackage{array}
\usepackage[caption=false,font=normalsize,labelfont=sf,textfont=sf]{subfig}
\usepackage{textcomp}
\usepackage{stfloats}
\usepackage{url}
\usepackage{verbatim}
\usepackage{graphicx}
\usepackage{booktabs,multirow}
\usepackage{makecell}
\usepackage{verbatim}

\begin{document}

\title{Towards Data-centric Machine Learning on Directed Graphs: a Survey}

\author{Henan Sun}
\authornote{Both authors contributed equally to this research.}
\email{magneto0617@foxmail.com}
\author{Xunkai Li}
\authornotemark[1]
\email{cs.xunkai.li@gmail.com}
\author{Daohan Su}
\email{dhsu@bit.edu.cn}
\affiliation{%
  \institution{Beijing Institute of Technology}
  \city{Beijing}
  \country{China}
}

\author{Junyi Han}
\email{hanjy5521@mails.jlu.edu.cn}
\affiliation{%
  \institution{Jilin University}
  \city{Jilin}
  \country{China}
}

\author{Rong-Hua Li}
\email{lironghuabit@126.com}
\author{Guoren Wang}
\email{wanggrbit@gmail.com}
\affiliation{%
  \institution{Beijing Institute of Technology}
  \city{Beijing}
  \country{China}
}

\renewcommand{\shortauthors}{Sun et al.}

\begin{abstract}
    In recent years, Graph Neural Networks (GNNs) have made significant advances in processing structured data. 
    However, most of them primarily adopted a model-centric approach, which simplifies graphs by converting them into undirected formats and emphasizes model designs.
    This approach is inherently limited in real-world applications due to the unavoidable information loss in simple undirected graphs and the model optimization challenges that arise when exceeding the upper bounds of this sub-optimal data representational capacity.
    As a result, there has been a shift toward data-centric methods that prioritize improving graph quality and representation.
    Specifically, various types of graphs can be derived from naturally structured data, including heterogeneous graphs, hypergraphs, and directed graphs. 
    Among these, directed graphs offer distinct advantages in topological systems by modeling causal relationships, and directed GNNs have been extensively studied in recent years.
    However, a comprehensive survey of this emerging topic is still lacking.
    Therefore, we aim to provide a comprehensive review of directed graph learning, with a particular focus on a data-centric perspective.
    Specifically, we first introduce a novel taxonomy for existing studies.
    Subsequently, we re-examine these methods from the data-centric perspective, with an emphasis on understanding and improving data representation.
    It demonstrates that a deep understanding of directed graphs and their quality plays a crucial role in model performance.
    Additionally, we explore the diverse applications of directed GNNs across 10+ domains, highlighting their broad applicability. 
    Finally, we identify key opportunities and challenges within the field, offering insights that can guide future research and development in directed graph learning.
\end{abstract}

\ccsdesc[500]{Computing methodologies~Artificial intelligence}

\keywords{Graph neural networks, directed GNN, data-centric graph learning}

\maketitle

\section{Introduction}
\label{sec:intro}
    Graph Neural Networks (GNNs), as an emerging paradigm in machine learning for structured data, have attracted significant attention within the computing community~\cite{wu2020gnn_survey1, zhou2022gnn_survey2, bessadok2022gnn_survey3, song2022gnn_survey4}. 
    Specifically, GNNs overcome the limitations of traditional methods by employing recursive message-passing mechanisms that facilitate information flow across the graph structure.
    Through this iterative process, GNNs generate rich, embedded representations of nodes, edges, and entire graphs for various graph-based downstream tasks, effectively capturing the underlying structural and relational information. 
    Their wide applicability across industrial domains, such as recommendation systems~\cite{zhao2021ugrec_directed_app_recommendation1, virinchi2022_directed_app_recommendation2, tiady2024merlin_directed_app_recommendation3, Virinchi2023_directed_app_recommendation4, liu2024exploring_directed_app_recommendation5}, anomaly detection~\cite{tang2022app_detection1, chen2022app_detection2, duan2023app_detection3}, traffic flows~\cite{chen2021temporal_directed_app_tf1, wang2022traffic_directed_app_tf2, li2022spatio_directed_app_tf3, xie2019sequential_directed_app_tf4, sharma2023graph_directed_app_tf5} and drug-drug interaction networks~\cite{voitsitskyi20233dprotdta_directed_app_ddi1, ma2023dual_directed_app_ddi2, yan2024predicting_directed_app_ddi3, li2024deepdrug_directed_app_ddi4, huang2024structure_directed_app_ddi5}.

    In the development of GNNs, significant attention has been devoted to \textbf{model-centric} studies, which focus on designing complex GNNs tailored to specific datasets and diverse learning tasks.
    From the data perspective, researchers often simplify graph data by converting it into undirected forms, enabling a focus on optimizing model architectures. 
    These approaches, commonly referred to as undirected GNNs, aim to enhance GNN performance by prioritizing model innovation over complex data representations, driving progress in various application domains~\cite{kipf2016gcn, hamilton2017graphsage, velivckovic2017gat, wu2019sgc}. 
    However, this simplification assumes that the input data has been adequately preprocessed or refined into a suitable format—an assumption that often does not hold in most real-world scenarios~\cite{whang2023_data_centric_ml1}. 
    Consequently, the limitations of undirected GNNs hinder their practical effectiveness in real-world applications.
    
    \textit{Limitation 1: Hindering Advancement in Academia.}
    From a theoretical neural representation perspective, simplifying natural graphs into undirected forms overlooks crucial dimensions of information, such as node/edge heterogeneity (heterogeneous graphs), group relationships between nodes and edges (hypergraphs), and edge directionality (directed graphs). 
    The loss of such critical information cannot be recovered during the training phase of any well-designed model. 
    Once essential data characteristics, like directed relationships or node heterogeneity, are omitted in the graph construction stage, the model's learning process is inherently limited. 
    This constraint prevents the model from fully capturing the complexity of the original graph structure, thereby hindering its ability to make accurate predictions or derive meaningful insights from incomplete data representations.
    In other words, sub-optimal data representations create a model optimization dilemma, where complex neural architectures may fall into local minima, overlooking the crucial role of data representation in dictating the upper bound of the model's representational capacity.
    For instance, ignoring edge directionality in directed graphs may lead to hindering models' capability of addressing the entanglement of homophily and heterophily (whether connected nodes share similar feature or labels) that has long plagued the graph learning community in the context of undirected graphs~\cite{dirgnn_rossi_2023, sun2023adpa}.
    
    \textit{Limitation 2: Hindering Deployment in Industry.}
    From an industrial deployment perspective, simplified undirected graph representations often fall short of capturing the complex, nuanced relationships that exist in real-world applications. 
    While undirected graphs can offer a simplified view, they fail to account for critical heterogeneous, high-order, and directed information that is essential in many industrial applications. 
    In particular, the lack of directionality in undirected graph representations limits the model's ability to accurately reflect the asymmetrical relationships that are often fundamental to understanding the underlying system of industrial applications.
    For instance, in drug-drug interaction networks~\cite{voitsitskyi20233dprotdta_directed_app_ddi1, ma2023dual_directed_app_ddi2, yan2024predicting_directed_app_ddi3, li2024deepdrug_directed_app_ddi4, huang2024structure_directed_app_ddi5}, modeling the graph as directed is crucial due to the differing roles that drugs play within the network. 
    Each drug in the network can interact with other drugs in a specific, directed manner, with one drug potentially influencing or modifying the effect of another. 
    The absence of this directed information not only impedes the model’s predictive accuracy but also limits its practical applicability in critical downstream fields of drug-drug interaction networks, such as healthcare where precise, data-driven decision-making is paramount. A similar challenge is evident across various industrial application domains, ranging from phishing detection efforts~\cite{huang2024peae_directed_app_pd1, ratra2024graph_directed_app_pd2, zhang2024grabphisher_directed_app_pd3, wang2021tsgn_directed_app_pd4, kim2023graph_directed_app_pd5} to brain network analysis~\cite{alaei2023_directed_app_bna1, kong2022causal_directed_app_bna2, park2023convolving_directed_app_bna3, yu2022learning_directed_app_bna4, cao2024dementia_directed_app_bna5}. These domains face similar dilemmas, highlighting the broader implications of these limitations within real-world applications of graph machine learning.

\begin{figure*}[t]
  \includegraphics[width=\textwidth]{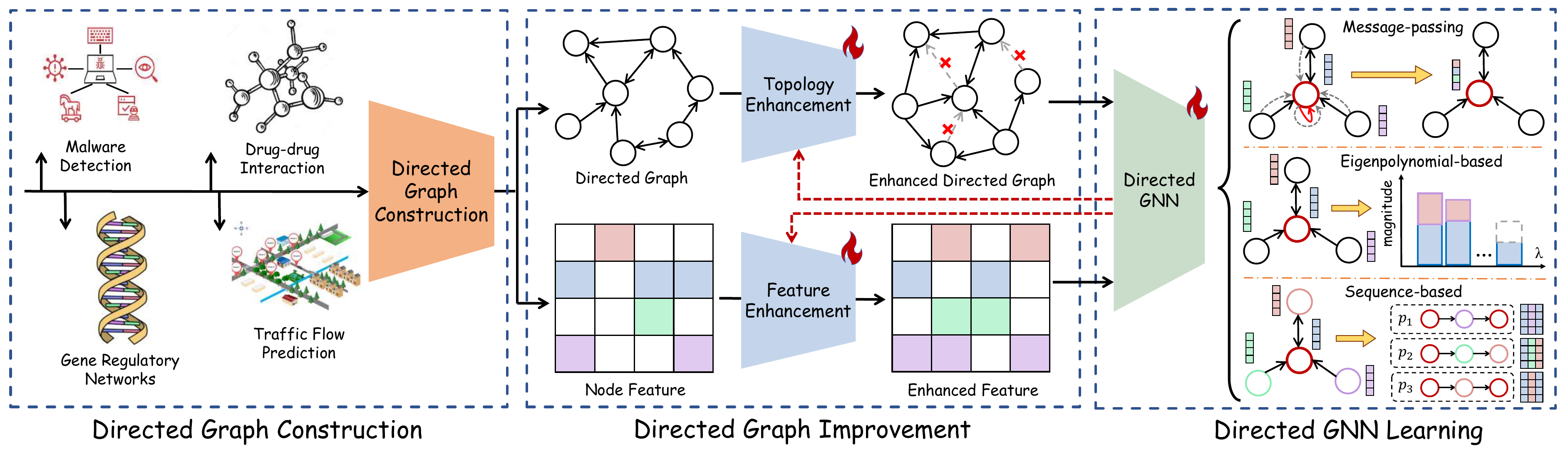}
  \captionsetup{font={small,stretch=1}}
  \caption{The pipeline of data-centric graph machine learning.
   }
  \label{fig: data-centric}
  \vspace{-0.2cm}
\end{figure*}

    The limitations of undirected GNNs discussed earlier highlight the inherent challenges of model-centric approaches to graph learning, emphasizing the need for a \textbf{data-centric} paradigm. As illustrated in Fig.~\ref{fig: data-centric}, the data-centric graph learning pipeline typically consists of three key stages: graph construction, graph improvement, and GNN learning.
    In the first stage, raw data from diverse application domains—such as malware detection, drug-drug interaction analysis, gene regulatory network modeling, and traffic flow prediction—are systematically transformed into graph structures. The second stage involves refining these graphs through various enhancement techniques, including topological improvements and feature augmentation, to produce higher-quality graph data that is better suited for training. Finally, different types of GNNs, such as those based on message-passing, eigenpolynomial, and sequence-based framework, are employed to derive effective models tailored for specific downstream tasks. This structured approach underscores the critical role of data quality and preprocessing in achieving robust and reliable graph learning outcomes.
    
    Naturally, from the data-centric perspective, graph data can be effectively structured into various graph types, such as heterogeneous graphs, hypergraphs, and directed graphs~\cite{whang2023_data_centric_ml1, yang2023_data_centric_ml2, zheng2023towards_data_centric_ml3}. 
    Each of these types facilitates the modeling of unique structural and relational characteristics inherent in the data, which undirected GNNs fail to capture.
    Among these, heterogeneous graphs and hypergraphs have gained significant attention in recent years, sparking a wave of research interest.
    Specifically, unlike undirected graphs, heterogeneous graphs consist of different types of nodes and edges, with each node potentially belonging to distinct label classes.
    This classification leads to separate feature spaces for different types of nodes, requiring the development of specialized models to understand the relationships between these varied nodes. 
    Such models can effectively capture diverse structural patterns, but they often come with the trade-offs of increased complexity and higher computational demands~\cite{wang2022_heteroGNN_survey, bing2023_heteroGNN_survey, yang2020_heteroGNN_survey}.
    Hypergraphs, an extension of simple graphs, introduce the concept of hyperedges, which can connect an arbitrary number of nodes, allowing for the representation of more complex, multidimensional relationships. 
    This added flexibility significantly enhances the structural representation capabilities compared to undirected GNNs. 
    However, the ability of hyperedges to connect multiple nodes requires more extensive message-passing processes, leading to increased computational complexity~\cite{kim2024_HNN_survey, wang2022_HNN_survey, antelmi2023_HNN_survey}.
    
    In conclusion, while data-centric approaches like heterogeneous graphs and hypergraphs offer more comprehensive representations of graph data, they also present challenges related to model complexity and computational efficiency.
    These challenges must be carefully addressed in order to fully leverage the potential of these graph types in real-world applications.
    In recent years, directed graphs also have garnered significant attention due to their ability to model complex causal relationships within topological systems. 
    However, there remains a lack of comprehensive surveys on this topic, which has inspired the motivation for this paper.

    As previously mentioned, several surveys have explored heterogeneous GNNs~\cite{wang2022_heteroGNN_survey, bing2023_heteroGNN_survey, yang2020_heteroGNN_survey} and Hypergraph Neural Networks (HNNs)~\cite{kim2024_HNN_survey, wang2022_HNN_survey, antelmi2023_HNN_survey} from a data-centric perspective. 
    However, directed GNNs, a crucial area of study, have not received similar attention in the form of dedicated surveys. 
    This gap highlights an underserved aspect of current research, suggesting that directed GNNs warrant further exploration to fully understand their potential applications and implications within data-centric frameworks. 
    Specifically, the detailed importance are as follow:

\noindent
    \textit{(1) Unique data understanding in academic}: 
    Unlike undirected GNNs, directed GNNs preserve the directionality of edges, which not only facilitates a clearer understanding of graph data but also enhances the model's capacity to capture directed relationships and the broader semantic context they convey. This preservation of edge directionality improves the theoretical upper bound of the representational power of graph neural networks. Furthermore, directed GNNs possess the flexibility to process both directed and undirected graphs, expanding their applicability in diverse scenarios. For example, the presence of edge directionality enables directed GNNs to leverage various graph views—such as the graph topological view, spectral view, and sequence view—to better interpret the original graph data. This results in the design of more robust models capable of capturing the asymmetric relationships between node pairs. Additionally, the inclusion of directed information allows for the development of advanced graph encoding mechanisms, such as dual encoding~\cite{kollias2022nste, he2022dimpa, wangdirected} and magnetic laplacian encoding~\cite{zhang2021magnet,lin2023framelet_gnn, li2024lightdic}, which significantly enhance the model's ability to represent graph structures and improve its performance.
    This advantage positions directed GNNs as a practical alternative within the broader landscape of graph learning techniques, enabling their use across diverse domains.

\noindent
    \textit{(2) Widespread industrial application}: 
    In contrast to undirected GNNs, directed GNNs offer a broader range of real-world applications due to their superior ability to model causal relationships within complex topological systems, such as recommendation systems~\cite{zhao2021ugrec_directed_app_recommendation1, virinchi2022_directed_app_recommendation2, tiady2024merlin_directed_app_recommendation3, Virinchi2023_directed_app_recommendation4, liu2024exploring_directed_app_recommendation5}, conversational emotion recognition~\cite{gan2024graph_directed_app_er1, ghosal2019dialoguegcn_directed_app_er2, shen2021_directed_app_er3, li2023graphcfc_directed_app_er4, joshi2022cogmen_directed_app_er5}, traffic flow prediction~\cite{chen2021temporal_directed_app_tf1, wang2022traffic_directed_app_tf2, li2022spatio_directed_app_tf3, xie2019sequential_directed_app_tf4, sharma2023graph_directed_app_tf5}, drug-drug interaction networks~\cite{voitsitskyi20233dprotdta_directed_app_ddi1, ma2023dual_directed_app_ddi2, yan2024predicting_directed_app_ddi3, li2024deepdrug_directed_app_ddi4, huang2024structure_directed_app_ddi5}, pedestrian trajectory forecasting~\cite{su2022trajectory_directed_app_ptf1, wang2023trajectory_directed_app_ptf2, zhang2023dual_directed_app_ptf3, liu2022mdst_directed_app_ptf4, shi2021sgcn_directed_app_ptf5}, gene regulatory networks~\cite{wei2024inference_directed_app_grn1, jereesh2024reconstruction_directed_app_grn2, mao2023predicting_directed_app_grn3, feng2023gene_directed_app_grn4, chen2022graph_directed_app_grn5}, malware detection~\cite{zapzalka2024semantics_directed_app_md1, esmaeili2023gnn_directed_app_md2, busch2021nf_directed_app_md3, liu2023nt_directed_app_md4, 9200767liu2023nt_directed_app_md5}, brain network analysis~\cite{alaei2023_directed_app_bna1, kong2022causal_directed_app_bna2, park2023convolving_directed_app_bna3, yu2022learning_directed_app_bna4, cao2024dementia_directed_app_bna5}, social network analysis~\cite{bian2020_directed_app_social1, schweimer2022_directed_app_social2, song2021rumor_directed_app_social3, wei2021towards_directed_app_social4, yang2023rosgas_directed_app_social5}, financial network analysis~\cite{cheng2021risk_directed_app_financial1, wu2022grande_directed_app_financial2, egressy2024provably_directed_app_financial3, dileo2024discrete_directed_app_financial7, bukhori2023inductive_directed_app_financial8}, phishing detection~\cite{huang2024peae_directed_app_pd1, ratra2024graph_directed_app_pd2, zhang2024grabphisher_directed_app_pd3, wang2021tsgn_directed_app_pd4, kim2023graph_directed_app_pd5} and chip design~\cite{ji2022gnn_directed_app_chip1, ji2023gat_directed_app_chip2, alrahis2022gnn4rel_directed_app_chip3, ji2024gnn_directed_app_chip4, wang2024gns_directed_app_chip5}, as shown in Table~\ref{tab:DGNN_app}. 
    For instance, in Gene Regulatory Networks (GRNs)~\cite{wei2024inference_directed_app_grn1, jereesh2024reconstruction_directed_app_grn2, mao2023predicting_directed_app_grn3, feng2023gene_directed_app_grn4, chen2022graph_directed_app_grn5}, modeling these networks as directed graphs is crucial due to the distinct roles that genes play in regulating one another during protein synthesis.
    The directionality of interactions among genes must be preserved, as neglecting this aspect may result in the loss of essential regulatory relationships. 
    Similarly, in traffic flow prediction~\cite{chen2021temporal_directed_app_tf1, wang2022traffic_directed_app_tf2, li2022spatio_directed_app_tf3, xie2019sequential_directed_app_tf4, sharma2023graph_directed_app_tf5}, the inherent direction of vehicle movement necessitates representing these systems as directed graphs, as vehicles naturally move towards specific destinations. 
    Misrepresenting traffic flow networks using other graph types, such as undirected graphs, can lead to a fundamental misunderstanding of the underlying dynamics of traffic patterns. 
    Thus, adopting directed GNNs in these domains is essential for accurately capturing the complexities of interactions and improving model performance.

\begin{table*}[t]
\centering
\caption{The widespread applications of directed GNNs.}
\label{tab:DGNN_app}
\resizebox{\textwidth}{!}{
\begin{tabular}{@{}ccc@{}}
\toprule
Application Domains        &    Task Classification        & Models                                            \\ \midrule
Phishing Detection         &   Node Level         &   \makecell[c]{ GAE\_PDNA~\cite{ratra2024graph_directed_app_pd2}, GrabPhisher~\cite{zhang2024grabphisher_directed_app_pd3}, SIEGE~\cite{li2023siege_directed_app_pd6},
BM-GCN~\cite{zhang2023phishing_directed_app_pd7}, TransWalk~\cite{xiong2023ethereum_directed_app_pd8}, GL4AML~\cite{karim2024scalable_directed_app_pd9}}     \\
Traffic Flow Prediction    &   Node Level         & \makecell[c]{T-DGCN~\cite{chen2021temporal_directed_app_tf1}, Traffic-GGNN~\cite{wang2022traffic_directed_app_tf2}, STGNN~\cite{li2022spatio_directed_app_tf3},  SeqGNN~\cite{xie2019sequential_directed_app_tf4}, DGNNs~\cite{sharma2023graph_directed_app_tf5}, STGC-GNNs~\cite{he2023stgc_directed_app_tf6}, AGP-GNN~\cite{guo2023adaptive_directed_app_tf7}, TF-MVGNN~\cite{cheng2024tf_directed_app_tf8}}     \\
Recommendation System      &   Node Level         & \makecell[c]{UGRec~\cite{zhao2021ugrec_directed_app_recommendation1}, DAEMON~\cite{virinchi2022_directed_app_recommendation2}, MERLIN~\cite{tiady2024merlin_directed_app_recommendation3},  BLADE~\cite{Virinchi2023_directed_app_recommendation4}, EIIGNN~\cite{liu2024exploring_directed_app_recommendation5}, GNN-SNR~\cite{yu2022graph_directed_app_recommendation6}}            \\
Pedestrian Trajectory Forecasting        &   Node Level         & \makecell[c]{TFDGCNN~\cite{su2022trajectory_directed_app_ptf1}, TDAGCN~\cite{wang2023trajectory_directed_app_ptf2}, DBSTGNN~\cite{zhang2023dual_directed_app_ptf3}, MDST-DGCN~\cite{liu2022mdst_directed_app_ptf4}, SGCN~\cite{shi2021sgcn_directed_app_ptf5}, MTP-GAN~\cite{mo2022multi_directed_app_ptf6},
HSTGA~\cite{alghodhaifi2023holistic_directed_app_ptf7}, XHGP~\cite{limeros2023towards_directed_app_ptf8}}          \\
Conversational Emotion Recognition        &   Node Level         & \makecell[c]{GNN4CER~\cite{gan2024graph_directed_app_er1}, DialogueGCN~\cite{ghosal2019dialoguegcn_directed_app_er2}, DAG-ERC~\cite{shen2021_directed_app_er3}, GraphCFC~\cite{li2023graphcfc_directed_app_er4}, COGMEN~\cite{joshi2022cogmen_directed_app_er5}, GA2MIF~\cite{li2023ga2mif_directed_app_er6}, ProtoDAG~\cite{kang2023directed_directed_app_er7}, CER-GNN~\cite{lian2020conversational_directed_app_er8}}          \\
Financial Network Analysis &   Edge Level         & \makecell[c]{DGANN~\cite{cheng2021risk_directed_app_financial1}, GRANDE~\cite{wu2022grande_directed_app_financial2}, GNNBTT~\cite{li2023graph_directed_app_financial6},  T3GNN~\cite{dileo2024discrete_directed_app_financial7}, ILPG~\cite{bukhori2023inductive_directed_app_financial8}} \\
Gene Regulatory Networks   &   Edge Level         & \makecell[c]{DGCGRN~\cite{wei2024inference_directed_app_grn1}, GRN-IUGNN~\cite{jereesh2024reconstruction_directed_app_grn2}, GNNLink~\cite{mao2023predicting_directed_app_grn3},  GRINCD~\cite{feng2023gene_directed_app_grn4}, GENELink~\cite{chen2022graph_directed_app_grn5}, NIMCE~\cite{feng2020nimce_directed_app_grn6}}       \\
Chip Design                &   Graph Level         & \makecell[c]{GCPRMM~\cite{ji2022gnn_directed_app_chip1}, GACPRMM~\cite{ji2023gat_directed_app_chip2}, GNN4REL~\cite{alrahis2022gnn4rel_directed_app_chip3},  GCPVIFR~\cite{ji2024gnn_directed_app_chip4}, GNS~\cite{wang2024gns_directed_app_chip5}}            \\
Malware Detection      &   Graph Level         & \makecell[c]{SPNGNN~\cite{zapzalka2024semantics_directed_app_md1}, GNN-MDF~\cite{esmaeili2023gnn_directed_app_md2}, NF-GNN~\cite{busch2021nf_directed_app_md3}, NT-GNN~\cite{liu2023nt_directed_app_md4}, SDGNet~\cite{9200767liu2023nt_directed_app_md5}, MDGraph~\cite{li2024mdgraph_app_md6}, DawnGNNz~\cite{feng2024dawngnn_app_md7}, IMD-GCN~\cite{li2022intelligent_app_md8}}        \\
Drug-drug Interaction      &   Graph Level         & \makecell[c]{3DProtDTA~\cite{voitsitskyi20233dprotdta_directed_app_ddi1}, DGNN-DDI~\cite{ma2023dual_directed_app_ddi2}, GMIA~\cite{yan2024predicting_directed_app_ddi3},  DeepDrug~\cite{li2024deepdrug_directed_app_ddi4}, SISDTA~\cite{huang2024structure_directed_app_ddi5}, GMPNN~\cite{nyamabo2022drug_directed_app_ddi10}}        \\
Social Network Analysis    &   Graph Level         & \makecell[c]{Bi-GCN~\cite{bian2020_directed_app_social1}, IS-GNN~\cite{schweimer2022_directed_app_social2}, OID-GCN~\cite{song2021rumor_directed_app_social3},  EBGCN~\cite{wei2021towards_directed_app_social4}, RoSGAS~\cite{yang2023rosgas_directed_app_social5} }            \\
Brain Network Analysis     &   Graph Level         & \makecell[c]{DBNA~\cite{alaei2023_directed_app_bna1}, CGCNN~\cite{kong2022causal_directed_app_bna2}, Hodge-GNN~\cite{park2023convolving_directed_app_bna3},  TBDS~\cite{yu2022learning_directed_app_bna4}, DSL-GNN~\cite{cao2024dementia_directed_app_bna5}, DABNet~\cite{yu2023deep_directed_app_bna6}, TSGBF~\cite{wang2023eeg_directed_app_bna7}, SSGNN4ESA~\cite{tang2021self_directed_app_bna8}}             \\ \bottomrule
\end{tabular}}
\end{table*}

As previously discussed, several surveys have already explored heterogeneous GNNs~\cite{wang2022_heteroGNN_survey, bing2023_heteroGNN_survey, yang2020_heteroGNN_survey} and HNNs~\cite{kim2024_HNN_survey, wang2022_HNN_survey, antelmi2023_HNN_survey}. However, to the best of our knowledge, no survey has yet specifically addressed directed GNNs, despite their distinct advantages in both academic research and industrial applications. In light of this gap, this survey seeks to fill the void by providing a comprehensive overview about directed GNNs, examining their unique theoretical characteristics and contributions in the industrial applications. Our goal is to advance the understandings of directed GNNs and contribute to the ongoing development of data-centric directed graph learning. 

In conclusion, the primary contributions of this work are summarized as follows:

\begin{itemize}
    \item \textbf{Comprehensive Review.} 
    To the best of our knowledge, this study is the first comprehensive survey of data-centric machine learning applied to directed graphs. 
    In this survey, we provide a thorough examination of the current landscape and future prospects of data-centric directed graph machine learning, offering an in-depth review of recent advancements in graph neural networks research. 
    This work aims to deepen the understanding and practical application of directed GNNs while fostering future research directions in this emerging field. 
    By highlighting critical developments and identifying potential areas for further exploration, we seek to contribute to the ongoing discourse on directed graph machine learning methodologies and their applications.

    \item \textbf{New Taxonomies.} 
     In this survey, we present a novel taxonomy for the classification of directed GNNs, which encompasses three principal frameworks: the message-passing framework, the eigenpolynomial-based framework, and the sequence-based framework, as outlined in Table\ref{tab:taxnomy}. This categorization aims to enhance the understanding of the varied methodologies underlying directed GNNs and to facilitate the selection of appropriate techniques for specific application scenarios. By rigorously defining these categories, the proposed taxonomy seeks to advance the discourse on machine learning for directed graphs, offering a structured foundation for future research endeavors and practical implementations in this rapidly evolving field.

    \item \textbf{Data-centric Revisiting.} 
    We revisit the existing directed GNNs from a data-centric perspective, emphasizing how these models interpret and refine the graph data to enhance predictive accuracy and overall effectiveness.
    Specifically, we introduce three key perspectives for examining graph data: the graph topology view, the graph spectral view, and the graph sequence view. 
    Building upon these perspectives, we then explore various techniques for improving the quality of graph data, including topological enhancement and node feature enhancement.
    These distinct views and advanced enhancement strategies are vital for improving the theoretical expressiveness and performance of directed GNN models, offering valuable insights to further promote advance research and applications in directed graph learning.

    \item \textbf{Industrial Application Overview.} This study presents a comprehensive, data-centric examination of the diverse industrial applications of directed GNNs, spanning domains such as phishing detection~\cite{huang2024peae_directed_app_pd1, ratra2024graph_directed_app_pd2, zhang2024grabphisher_directed_app_pd3, wang2021tsgn_directed_app_pd4, kim2023graph_directed_app_pd5} and brain network analysis~\cite{alaei2023_directed_app_bna1, kong2022causal_directed_app_bna2, park2023convolving_directed_app_bna3, yu2022learning_directed_app_bna4, cao2024dementia_directed_app_bna5}, as summarized in Table~\ref{tab:DGNN_app}. These applications are revisited through a multi-level graph task lens, encompassing node-level, edge-level, and graph-level tasks. This approach provides several critical insights into the strengths and challenges of directed GNNs, contributing to their potential optimization and deployment in real-world industrial scenarios.

    \item \textbf{Future Directions.} 
    We critically evaluate the limitations of current state-of-the-art directed GNNs from the perspectives of theoretical analysis and industrial applications, and propose several potential research directions for future exploration. 
    Our analysis addresses key aspects about data-centric graph machine learning, such as problem formulation, graph construction, modeling techniques, and scalability challenges.
    In light of these, we aim to enhance the efficiency and applicability of directed GNNs in tackling complex real-world problems. 
    This survey serves to inform future investigations focused on advancing directed GNN methodologies.
\end{itemize}

\textbf{Organization.} 
    The remainder of this survey is organized as follows: Section~\ref{sec:backgrounds} provides an overview of directed graph machine learning, including key notations and various downstream tasks, as outlined in Table~\ref{tab:notations}.
    Section~\ref{sec:dgnnm} introduces a novel taxonomy for directed GNNs and offers a comprehensive analysis of existing models within different frameworks, as shown in Table~\ref{tab:taxnomy}. 
    Section~\ref{sec:revisit} revisits directed GNNs from a data-centric perspective, highlighting the importance of both understanding and improving graph data quality to enhance model performance, as presented in Table~\ref{tab:dcv}. 
    Section~\ref{sec:applications} presents real-world application examples enabled by the directed GNNs discussed earlier.
    Section~\ref{sec:future} explores future research directions for data-centric directed GNNs, focusing on opportunities and open challenges.
    Finally, Section~\ref{sec:conclusion} concludes the survey.

\section{Backgrounds}
\label{sec:backgrounds}
This section outlines the foundational definitions for directed graph neural networks, abbreviated as directed GNNs. First, we introduce the notations utilized throughout this paper, summarized in Table~\ref{tab:notations}. Following that, we describe three kinds of common downstream tasks that are commonly adopted in the directed GNNs researches.

\subsection{Notations}
Through out this paper, we use bold uppercase characters, such as $\textbf{A}$, to denote matrices, bold lowercase characters (e.g. $\textbf{x}$) to denote vectors, and calligraphic characters, for example $\mathcal{V}$, to denote sets. We use Pytorch-style indexing conventions for matices and vectors. For instance, $\textbf{A}[i,j]$ denotes the value of matrix $\textbf{A}$ at the $i$-th row and the $j$-th column. Also, $\textbf{A}[i,:]$ and $\textbf{A}[:,j]$ denotes the $i$-th row and $j$-th column of matrix $\textbf{A}$, respectively, and $\textbf{x}[i]$ denotes the $i$-th value of vector $\textbf{x}$. The notations employed in this survey are summarized in Table~\ref{tab:notations}.

We consider a general graph representation method $G = (\mathcal{V}, \mathcal{E})$ with $|\mathcal{V}|=n$ nodes and $|\mathcal{E}|=m$ edges, in which $v_i, v_j \in \mathcal{V}$ denotes $i$-th and $j$-th nodes of graph $G$, and $e_{ij}=(v_i, v_j)\in \mathcal{V}$ denotes the directed edge from node $v_i$ to node $v_j$. We can also represent the graph structure by the adjacency matrix $\textbf{A}\in \left\{ 0,1 \right\}^{n\times n}$, where $\textbf{A}[i,j]=1$ represent $e_{ij}\in \mathcal{E}$ and $\textbf{A}[i,j]=0$ represent $e_{ij}\notin \mathcal{E}$. A graph may contain node attributes, effectively organized within a node feature matrix $\textbf{X}\in \mathbb{R}^{n\times f}$, where $\textbf{x}_i\in \mathbb{R}^f$ or $\textbf{X}[i,:]\in \mathbb{R}^f$ represent the feature of node $v_i$. A graph may also have node-level labels, represented by a label vector $\textbf{y}$, where the $i$-th value $\textbf{y}[i]$ denotes the label of node $v_i$.

\begin{table*}[h!t]
\centering
\caption{Summary of the Notations Used throughout This Survey.}
\label{tab:notations}
\resizebox{\textwidth}{!}{
\begin{tabular}{@{}ccc@{}}
\toprule
Category & Notation & Explanation \\ \midrule
\multirow{8}{*}{\makecell[c]{General}}        &    $G, \mathcal{V}, \mathcal{E}$       &  \makecell[c]{The graph, the node set and the edge set.}             \\
        &   $v_i, e_{ij}$        &  \makecell[c]{Node $v_i$ and the edge connecting node $v_i$ and node $v_j$, respectively.} \\
         &   $\textbf{A},\tilde{\textbf{A}}$       &  \makecell[c]{The adjacency matrix and the one with self-loop.}           \\
         &   $\textbf{D},\tilde{\textbf{D}}$       &  \makecell[c]{The degree matrix and the one with self-loop.}           \\
         &   $\textbf{L}, \tilde{\textbf{L}}$       & \makecell[c]{The Laplacian matrix and the one with self-loop.}          \\
         &  $\textbf{P}$    &  \makecell[c]{The random walk transition matrix.}         \\
         &   $\textbf{X}, \textbf{x}_i$       &   \makecell[c]{The feature matrix and the feature vector of user $v_i$.}          \\
         &    $\textbf{y}$      &    \makecell[c]{The label vector.}         \\ \cmidrule{1-3}
\multirow{5}{*}{\makecell[c]{Message-passing Framework}}         &    $Prop(\cdot), Agg(\cdot)$      &  \makecell[c]{The propagate and aggregate operators in message-passing framework.}           \\
         &   $\mathbf{\Pi}$       &    \makecell[c]{The propagation kernel.}         \\
         &   $\textbf{Z}_{u}^{(l)}$       &  \makecell[c]{The aggregated embedding of node $u$ at $l$-th layer.}           \\
         &   $\textbf{H}_u^{(l)}$       &   \makecell[c]{The propagated message from node $u$ at $l$-th layer.}          \\
         &   $\textbf{W}^{(l)}$       &   \makecell[c]{The learnable matrix at $l$-th layer.}           \\ \cmidrule{1-3}
\multirow{13.5}{*}{\makecell[c]{Eigenpolynomial-based \\Framework}}         &   $\textbf{g}_{\theta}$       &   \makecell[c]{The graph filter.}            \\
         &   $\textbf{U}, \mathbf{\Lambda}$       &  \makecell[c]{The eigenvector matrix and the eigenvalue matrix.}      \\
         &   $\textbf{T}_i(\cdot)$       &   \makecell[c]{The eigenpolynomial for the $i$-th eigenvalue.}         \\
         &   $\tilde{\mathbf{\Delta}}$       & \makecell[c]{The motif Laplacian matrix employed in the MotifNet~\cite{monti2018motifnet}.}              \\
         &   $\phi_{norm}, \mathbf{\Phi}$       &  \makecell[c]{The Perron vector from the Perron-Frobenius Theorem~\cite{horn2012matrix} and its diagonal matrix \\employed in the DGCN~\cite{tong2020dgcn}.}  \\
         &   $\textbf{L}^{sym}$       &  The Laplacian matrix employed in the DGCN~\cite{tong2020dgcn}. \\
         &  $\textbf{P}_{ppr}$      &  \makecell[c]{The personalized PageRank transition matrix employed in the DiGCN~\cite{tong2020digcn}.}       \\ 
         &  $\pi_{appr}, \mathbf{\Pi}_{appr}$        &  \makecell[c]{The approximation of the Perron vector of $\textbf{P}_{ppr}$ and its diagonal matrix employed in the DiGCN~\cite{tong2020digcn}.}         \\ 
         &  $\textbf{H}^{(q)}$        &  \makecell[c]{The magnetic Laplacian matrix employed in the MagNet~\cite{zhang2021magnet}.}\\
         &  $\textbf{T}_{LR}$    &  \makecell[c]{The LinearRank eigenpolynomial employed in the MGC~\cite{zhang2021mgc}.} \\
         &  $\mathcal{S}_p(\textbf{x})$    &  \makecell[c]{The Dirichlet energy of graph signals employed in the MGC~\cite{zhang2021mgc}.}         \\
         &  $\rho_{n,s}^{(q)}(m), \varrho_{n,s,r}^{(q)}(m)$    &   \makecell[c]{The framelets for the low-pass filter and the high-pass filter employed in the Framelet-MagNet~\cite{lin2023_framelet-magnet}.}\\        
         &  $\mathbf{\Psi}^{fwd/bwd}$    &  \makecell[c]{The forward and backward graph filter bases employed in the HoloNet~\cite{koke2023holonets}.}   \\ \cmidrule{1-3}
\multirow{4}{*}{\makecell[c]{Sequence-based Framework}}         & $\Upsilon(G), \Phi(G)$     &  \makecell[c]{The node tokenizer and the subgraph tokenizer of graph $G$.}    \\
         &   $\Psi_i$   &  \makecell[c]{The encoder of node $v_i$ for the node tokenizer or the subgraph tokenizer.}\\ 
         &  $\textbf{Q}, \textbf{K}, \textbf{V}$    &  \makecell[c]{The query, key, value matrices in the Graph Transformer~\cite{cai2020graph}.}         \\
         &  $N_k(v_i)$    &  \makecell[c]{The neighborhood within $k$-hops of node $v_i$.}         \\ \bottomrule
\end{tabular}}
\end{table*}

\subsection{Downstream Tasks}
Based on the scale of prediction targets, we classify the downstream tasks of directed GNNs into three categories: node-level tasks, edge-level tasks, and graph-level tasks.

\noindent
\textbf{Node-level Tasks.} Node-level directed GNNs tasks are designed to predict specific properties for each node within one or more graphs. Common node-level tasks include node classification~\cite{kipf2016gcn}, node regression~\cite{rong2020self_node_regression}, and node clustering~\cite{pan2018adversarially_node_clustering}. Node classification, the most widely studied of these tasks, involves predicting a discrete label for each node. In training models for node classification, a cross-entropy loss function is often applied to minimize the discrepancy between predicted logits and the true labels of training nodes. However, node regression seeks to predict continuous properties of each node, while node clustering aims to group nodes into distinct clusters without requiring access to label information.

\noindent
\textbf{Edge-level Tasks.} Edge-level directed GNNs tasks aim to predict specific properties of each edge within one or more graphs. Common edge-level tasks include edge classification~\cite{aggarwal2016_edge_classification} and link prediction~\cite{Zhang18link_prediction1, cai2021link_prediction2, link_prediction3}. For example, link prediction seeks to determine the likelihood and its directionality of a connection existing between two nodes. Typically, this task is framed as a ternary classification problem, with a cross-entropy loss function applied to optimize link prediction models. Unlike link prediction, which focuses on the existence and the directionality of edges, edge classification predicts discrete labels for edges, identifying its particular characteristics rather than solely its presence.

\noindent
\textbf{Graph-level Tasks.} In contrast to node- and edge-level tasks, graph-level tasks focus on modeling and making predictions at the level of the entire graph, rather than on nodes or edges. These tasks commonly include graph classification~\cite{zhang2019graph_classification1, ma2019graph_classification2, yang2022graph_classification3, vincent2022graph_classification4}, graph regression~\cite{rong2020self_graph_regression}, graph matching~\cite{bai2019simgnn_graph_matching}, and graph generation~\cite{liao2019efficient_graph_generation}. Among these, graph classification serves as a foundational graph-level task, involving the assignment of a label or a class to an entire graph. To optimize models for graph classification, the cross-entropy between graph-level predictions and true labels is minimized. Other graph-level tasks address distinct objectives: graph regression aims to predict continuous properties of various graphs, graph matching seeks to evaluate the degree of similarity between any given pairs of graphs, and graph generation focuses on creating new graph samples that are suitable for other graph-related tasks.

\section{Directed Graph Neural Networks Methods}
\label{sec:dgnnm}
Directed GNNs are designed to learn representations that embed directed graphs into a low-dimensional space, while aiming to retain the intrinsic properties of the original directed graphs. Directed GNN methods can be categorized into three primary frameworks: the message-passing framework, eigenpolynomial-based framework, and sequence-based framework, as shown in Table~\ref{tab:taxnomy}. The primary distinction among these three categories lies in the fundamental frameworks each employs for processing directed graph data, which are summarized as follow:

\begin{table*}[h!t]
\centering
\caption{Taxonomy and representative works of directed GNNs.}
\label{tab:taxnomy}
\resizebox{\textwidth}{!}{
\begin{tabular}{@{}clc@{}}
\toprule
Taxonomy                                         & Innovation Perspective    & Models \\ \midrule
\multirow{3.5}{*}{Message-passing Framework}       & (\S \ref{sec:mpf}) Propogator                &  \makecell[c]{EDGNN~\cite{wu2021fedgnn}, DGCN~\cite{tong2020dgcn}, GNNDM~\cite{egressy2024provably_directed_app_financial3}, LightDiC~\cite{li2024lightdic}}      \\  \cmidrule{2-3}
                                                 & (\S \ref{sec:mpf})  Aggregator                &  \makecell[c]{D-VAE~\cite{zhang2019dvae}, NSTE~\cite{kollias2022nste}, GRANDE~\cite{wu2022grande_directed_app_financial2}}      \\  \cmidrule{2-3}
                                                 & (\S \ref{sec:mpf}) Both                      &  \makecell[c]{DAGNN~\cite{thost2021dagnn}, D-HYPR~\cite{zhou2022dhypr}, DIMPA~\cite{he2022dimpa}, A2DUG~\cite{maekawa2023a2dug}, Dir-GNN~\cite{dirgnn_rossi_2023}, ADPA~\cite{sun2023adpa}}      \\ \cmidrule{1-3}
\multirow{3.8}{*}{\makecell[c]{Eigenpolynomial-based \\
                                    Framework}} & (\S \ref{sec:ebf}) Graph Filter Basis        &  \makecell[c]{MotifNet~\cite{monti2018motifnet}, DGCN~\cite{ma2019spectral}, MagNet~\cite{zhang2021magnet}, DiGCL~\cite{tong2021directed}, Framelet-MagNet~\cite{lin2023framelet_gnn}}      \\  \cmidrule{2-3}
                                                 & (\S \ref{sec:ebf}) Eigenpolynomial           &  \makecell[c]{MGC~\cite{zhang2021mgc}}      \\  \cmidrule{2-3}
                                                 & (\S \ref{sec:ebf}) Both                      &  \makecell[c]{FDGCN~\cite{li2020scalable}, DiGCN~\cite{tong2020digcn}, HoloNet~\cite{koke2023holonets}}      \\ \cmidrule{1-3}
\multirow{3.5}{*}{Sequence-based Framework}        & (\S \ref{sec:sbf}) Node Token                & \makecell[c]{PACE~\cite{dong2022pace}, DAGRA~\cite{luo2024dagformer}, ADR-GNN~\cite{eliasof2023adr}, DiGT~\cite{wangdirected}, MMLPE~\cite{huang2024good}}       \\  \cmidrule{2-3}
                                                 & (\S \ref{sec:sbf}) Subgraph Token            &  \makecell[c]{DiRW~\cite{su2024dirw}}      \\  \cmidrule{2-3}
                                                 & (\S \ref{sec:sbf}) Both                      & \makecell[c]{TMDG~\cite{geisler2023transformers_meet_digraph}}       \\ \bottomrule
\end{tabular}}
\end{table*}

\begin{itemize}
    \item \textbf{Message-passing Framework.} The message-passing framework is the most widely employed approach in the design of directed GNNs. To generate node embeddings for directed graphs with asymmetric topologies, these models adapt message-passing mechanisms initially developed for undirected graph scenarios. More concretely, the framework typically involves two core operators: propagate operator $Prop(\cdot)$ and message aggregator $Agg(\cdot)$. The propagate operator facilitates the dissemination of messages throughout the graph topology, while the aggregator compiles messages from various neighboring nodes. Unlike undirected settings, however, both operators incorporate edge directionality, enabling them to operate effectively within directed graph structures.

    \item \textbf{Eigenpolynomial-based Framework.}
    The eigenpolynomial-based framework originates from the graph spectral theory. Following the Graph Fourier Transform, the filter operation on the graph spectrums can be equivalently formulated as a linear combination of eigenpolynomial. More concretely, this framework generally consists of two components: a filter basis and a linear polynomial. The filter basis is responsible for various kinds of graph eigen decomposition, such as $\textbf{L}=\textbf{D}-\textbf{A}$ derived from the Laplacian matrix, while the linear polynomial applies graph spectrum filtering, such as the low-passing filter via the Chebyshev polynomial. In contrast to undirected graphs, where the filter basis is typically real-symmetric, directed GNN methods within the eigenpolynomial-based framework require filter bases and linear polynomials designed specifically for asymmetric structures. These approaches enable effective operation within directed graph architectures.
    
    \item \textbf{Sequence-based Framework.} The sequence-based framework has gained significant recognition, particularly following the introduction of the Transformer architecture with its Encoder-Decoder framework, which highlights the importance of capturing long-range dependencies between item pairs in a sequence. Unlike GNNs based on the message-passing or eigenpolynomial-based frameworks, sequence-based GNNs extend beyond the inherent limitations of graph topology and graph spectrums. This approach begins with a tokenization process that transforms the graph structure into a sequence format, such as a node sequence or a subgraph sequence. Following tokenization, a suitable encoder is employed to aggregate information from the sequence, thereby enabling the model to capture long-range dependencies between item pairs within the sequence. The resulting embeddings are subsequently applied to a range of downstream tasks within the GNN framework, supporting enhanced predictive performance across diverse applications. 
\end{itemize}

\subsection{Message-passing Framework}
\label{sec:mpf}
\textbf{Motivations.}
Message passing mechanisms are central to GNNs, designed to enable information exchange among nodes to learn features within the graph structure. This mechanism typically consists of two main steps: \textit{Propagation} and \textit{Aggregation}. In the propagation step, each node receives and transmits feature information from neighboring nodes, reflecting inter-node connectivity. In the aggregation step, each node combines the received information to update its own representation. This layered approach allows the model to progressively capture both the localized neighborhood structure and each node’s distinctive characteristics, resulting in node representations that incorporate information from both the node itself and its surrounding neighbors within certain hops.

\noindent
In the case of undirected graphs where the adjacency matrix $\mathbf{A}$ is symmetric and $\mathbf{D}$ is the degree matrix of $\mathbf{A}$, many GNNs adhere to strictly spatial symmetric message-passing mechanisms, leveraging the balance of information from both directions of each edge.
For node $u$, the $l$-th message-passing scheme can be expressed as follows:
\begin{equation}
    \begin{aligned}
        \label{eq: Prevalent Message Passing}
        &\;\;\;\;\;\;\;\;\;\;\;\;\;\;\;\;\;      \mathbf{Z}_{u}^{(l)}=\operatorname{Agg}\left(\mathbf{W}^{(l)},\mathbf{H}_{u}^{(l)}\right),\\
        &\mathbf{H}_{u}^{(l)}=\operatorname{Prop}\left(\boldsymbol{\Pi},\left\{\mathbf{Z}_u^{(l-1)},\left\{\mathbf{Z}_v^{(l-1)},\forall (v,u)\in\mathcal{E}\right\}\right\}\right),
    \end{aligned}
\end{equation}
where $\boldsymbol{\Pi}$ is the propagation kernel (e.g. $\tilde{\mathbf{D}}^{-\frac{1}{2}}\tilde{\mathbf{A}}\tilde{\mathbf{D}}^{-\frac{1}{2}}$), defining the rule for information flow between nodes. $\mathbf{Z}_{u}^{(l)}$ represents the feature representation of node $u$ at the $l$-th layer of the model and $\mathbf{Z}_{u}^{(0)}=\mathbf{X}_{u}$.
The \textit{Propagation} function $\operatorname{Prop}\left(\cdot\right)$ gathers and distributes feature information from the neighbors based on the propagation rule $\boldsymbol{\Pi}$.
The \textit{Aggregation} function $\operatorname{Agg}\left(\cdot\right)$ 
is responsible for combining the feature information propagated from the neighboring nodes, which depends on the learnable parameter matrix $\mathbf{W}^{(l)}$.

\noindent
To capture the asymmetric topology of directed graphs, some spatial-based directed GNNs continue to follow the strict symmetric message-passing paradigm~\cite{huang2020cands, frasca2020sign, xu2018jknet, gamlp}.
However, the asymmetry of edges in directed graphs introduces directed dependencies between nodes, necessitating a message-passing process that considers edge directionality to accurately capture the topology of directed graphs.
Consequently, recent advances in directed GNNs have extended the traditional framework by designing customized directionality-aware \textit{Propagation} and \textit{Aggregation} mechanisms, thereby enhancing the model’s capacity to capture directed relationships within the graph.

\noindent
\textbf{Methods.}
EDGNN~\cite{jaume2019edgnn} extends traditional GNN propagation by separately handling incoming and outgoing edges and including edge labels. From the aggregation perspective, it applies different weight matrices to self-features and neighbor features, allowing the model to assign varying levels of importance to self-information and neighborhood information.
D-VAE~\cite{zhang2019dvae} employs an asynchronous message passing scheme to encode and decode DAGs, respecting the computation dependencies. The encoder and decoder of D-VAE both utilize GNNs to update the current node's state by aggregating the states of its predecessor nodes.
DGCN~\cite{tong2020dgcn} proposes a novel propagation mechanism that incorporates both first-order and second-order proximity of neighbors. This innovative approach leads to the development of aggregator that includes two distinct sets of independent learnable parameters.
DAGNN~\cite{thost2021dagnn} processes nodes based on the partial order relation of DAGs. In each layer, the node representation is updated by the attention-based propagator and the GRU-based aggregator.
NSTE~\cite{kollias2022nste} generates two vectors for each node, capturing its dual roles as both a source and a target in directed edges. The model introduces tunable parameters for weighting the degrees in its asymmetric propagation kernel.
 The use of separate aggregators for the source and target representations allows the model to distinguish the roles of nodes effectively.
GRANDE~\cite{wu2022grande_directed_app_financial2} leverages an attention mechanism to aggregate information from both in-neighbors and out-neighbors, effectively integrating neighborhood information from both incoming and outgoing edges of nodes. It also outputs both node and edge representations by performing message passing over the original graph and its augmented edge adjacency graph.
D-HYPR~\cite{zhou2022dhypr} propagates information through the graph by considering  multi-ordered four canonical types of partitioned neighborhoods. The model employs hyperbolic space for propagation and aggregation, which is particularly suited for modeling complex, non-Euclidean structures found in real-world digraphs.
DIMPA~\cite{he2022dimpa} distinctly propagates and aggregates the source and target nodes. It gathers neighborhood information across multiple hops within each layer to expand the model's receptive field. 
A2DUG~\cite{maekawa2023a2dug} utilizes the original adjacency matrix, the transpose of the adjacency matrix, and the undirected adjacency matrix as distinct propagation kernels to capture the nuances of both directed and undirected relationships within the graph. In the aggregation phase, the model leverages both propagated features and the adjacency lists, providing a comprehensive view of the local graph environment for each node.
Dir-GNN~\cite{dirgnn_rossi_2023} proposes performing separate propagation and aggregation for incoming and outgoing edges, with a learnable parameter that allows the model to weight the importance of in- and out-edges differently. The model also introduces a flexible propagation strategy that can be adapted to different GNN architectures.
ADPA~\cite{sun2023adpa} is distinguished by its adaptive exploration of appropriate neighborhood operators of directed $k$-order for propagation without the need for edge weights. It also incorporates a two-tiered system of node-adaptive attention mechanisms, which are designed to aggregate and optimize the representation of nodes within the graph.
GNNDM~\cite{egressy2024provably_directed_app_financial3} introduces a separate message-passing layer for handling incoming and outgoing edges, effectively adding reverse message passing to standard message passing framework. This adaptation transforms any standard GNN architecture into a directed GNN.
LightDiC~\cite{li2024lightdic} is an innovative and scalable adaptation of digraph convolution founded on the magnetic Laplacian. It employs the MGO as a novel propagation operator and pioneers the integration of complex numbers into the decoupling framework.

\subsection{Eigenpolynomial-based Framework}
\label{sec:ebf}
\textbf{Motivations.} The eigenpolynomial-based framework utilizes the Graph Fourier Transform~\cite{sandryhaila2013discrete} to facilitate the exchange of information between the spatial representation of graph topology and node features, and the spectral representation of eigenvectors and eigenvalues, thereby designing an appropriate spectral filter for graph representation learning. Initially, the framework decomposes the graph filter basis, which encapsulates information pertaining to graph topology and node features, into $n$ orthogonal eigenvectors along with their corresponding eigenvalues. This process effectively transforms information from the spatial domain into the spectral domain. Subsequently, graph filtering is applied to the eigenvalues to isolate the desired frequency components. Finally, an inverse Graph Fourier Transform is employed to revert the information from the spectral domain back to the spatial domain. According to the principles of Graph Fourier Transform theory, the filtering operation in the graph spectrum domain is equivalent to employing eigenpolynomials derived from the graph filter bases in the spatial domain. Consequently, methods utilizing the eigenpolynomial-based framework favor the implementation of graph representation learning through eigenpolynomials, thereby circumventing the computationally intensive process of decomposing eigenvectors. Formally, a graph spectrum filter can be expressed as follows:

\begin{equation}
    \textbf{x}*\textbf{g}_\theta = \textbf{U}\textbf{g}_{\theta}(\mathbf{\Lambda})\textbf{U}^T\textbf{x}=\sum_{i=1}^{K} \theta_i \textbf{T}_i(\tilde{\textbf{L}})\textbf{x},
\end{equation}

\noindent
where $\tilde{\textbf{L}}$ represents the graph filter basis, $\textbf{T}_i$ linearly combined with the coefficient $\theta_i$ represents the eigenpolynomial. Naturally, there are two main directions to improve the performance of methods in the eigenpolynomial-based framework: the graph filter basis and the eigenpolynomial.

\noindent
\textbf{Methods.} Conventional directed GNNs that adopt the eigenpolynomial-based framework typically focus on refining graph filter bases and eigenpolynomials within the real domain. This approach emphasizes leveraging real-valued spectral information to improve model design and representation capabilities, aligning with established methodologies in graph spectral theory for directed graphs.
MotifNet~\cite{monti2018motifnet} adapts the eigenpolynomial framework to the directed graph scenario by improving the graph filter basis with subgraph structures called motifs. it introduces the motif Laplacian $\tilde{\mathbf{\Delta}}=\textbf{I}-\tilde{\textbf{D}}_k^{-\frac{1}{2}}\tilde{\textbf{W}}_k\tilde{\textbf{D}}_k^{-\frac{1}{2}}$, where $\tilde{\textbf{W}}_k$ is the induced $k$-vertex motif adjacency matrix and $\tilde{\textbf{D}}_k$ is the corresponding degree matrix, to replace the original Laplacian matrix~\cite{kipf2016gcn} that is widely adopted in the undirected graph scenario, taking the directionality of edges into account. 
DGCN~\cite{ma2019spectral} enhances the graph filter basis with the Perron vector to facilitate graph spectrum filtering on the directed graphs. Since Perron-Frobenius Theorem~\cite{horn2012matrix} states that an irreducible matrix with non-negative entries has a unique left eigenvector with all entries positive, this vector can be found and normalized as the Perron vector $\mathbf{\phi}_{norm}$ with all values summed up to 1. Then, the improved graph filter basis is proposed: $\textbf{L}^{sum}=\textbf{I}-\frac{1}{2}(\mathbf{\Phi}^{\frac{1}{2}}\textbf{P}\mathbf{\Phi}^{-\frac{1}{2}}+\mathbf{\Phi}^{-\frac{1}{2}}\textbf{P}^T\mathbf{\Phi}^{\frac{1}{2}})$, where $\mathbf{\Phi}=diag(\mathbf{\phi}_{norm})$ and $\textbf{P}=\textbf{D}_{out}^{-1}\textbf{A}$.
FDGCN~\cite{li2020scalable} achieves the directed graph learning by simultaneously upgrading the graph filter basis and the eigenpolynomial. Similar to DGCN~\cite{ma2019spectral}, they also adopt the Perron vector to form their directed Laplacian filter, and approximate the eigenpolynomial with Chebyshev polynomial: $\textbf{x}*\textbf{g}_\theta \approx (\theta_0+\frac{2-\lambda_{max}}{\lambda_{max}}\theta_1)\textbf{Ix}-\frac{\theta_1}{\lambda_{max}}(\mathbf{\Phi}^{\frac{1}{2}}\textbf{P}\mathbf{\Phi}^{-\frac{1}{2}}+\mathbf{\Phi}^{-\frac{1}{2}}\textbf{P}^T\mathbf{\Phi}^{\frac{1}{2}})\textbf{x}$, where $\lambda_{max}$ is the greatest eigenvalue of the directed Laplacian matrix and $\theta_0=\frac{\lambda_{max}-2}{\lambda}\theta_1$. In order to save the time of finding the Perron vector $\mathbf{\phi}_{norm}$, they fix it as $(\frac{1}{n},\frac{1}{n},\cdots ,\frac{1}{n})^n$.
DiGCN~\cite{tong2020digcn} obtains a major improvement on the directed graph representation learning by further enhancing the graph basis and eigenpolynomial altogether. Different from DGCN~\cite{ma2019spectral} which adopts the random walk transition matrix $\textbf{P}=\textbf{D}_{out}^{-1}\textbf{A}$ as the basic component of the graph filter basis, DiGCN employs personalized PageRank transition matrix formulated as: $\textbf{P}_{ppr}=\left(\begin{smallmatrix} (1-\alpha)\tilde{\textbf{P}} & \alpha\textbf{1}^{n\times 1}\\ \frac{1}{n}\textbf{1}^{1\times n} & 0\end{smallmatrix}\right)$,
where $\tilde{\textbf{P}}=\tilde{\textbf{D}}^{-1}\tilde{\textbf{A}}, \tilde{\textbf{A}}=\textbf{A}+\textbf{I}$ and $\alpha$ is a hyper-parameter. According to Perron-Frobenius Theory~\cite{horn2012matrix}, there would be a Perron vector for $\textbf{P}_{ppr}$, denoted as $\mathbf{\pi}_{ppr}=(\mathbf{\pi}_{appr},\mathbf{\pi}_{\xi})$, where $\mathbf{\pi}_{appr}\in \mathbb{R}^n$ is the unique stationary distribution of the first n nodes and $\mathbf{\pi}_{\xi}\in \mathbb{R}^1$ is the unique stationary distribution of the auxiliary node $\xi$. They then approximate the personalized PageRank transition matrix and obtain the graph filter basis as $\textbf{L}_{appr}=\textbf{I}-\frac{1}{2}(\mathbf{\Pi}_{appr}^{\frac{1}{2}}\tilde{\textbf{P}}\mathbf{\Pi}_{appr}^{-\frac{1}{2}}+\mathbf{\Pi}_{appr}^{-\frac{1}{2}}\tilde{\textbf{P}}^{T}\mathbf{\Pi}_{appr}^{\frac{1}{2}})$, where $\mathbf{\Pi}_{appr}=\frac{1}{\Vert \pi_{appr} \Vert_1}diag(\pi_{appr})$. Meanwhile, they introduce $k^{th}$ order proximity to improve the eigenpolynimial, which focuses more on the node pair who share the common in- or out- neighbors, achieving larger reception field and learning multi-scale features in the directed graphs.
DiGCL~\cite{tong2021directed} improves the graph filter basis under the graph contrastive learning scenario to lower the time complexity of directed Laplacian matrix. They reduce the complexity of directed Laplacian operator of DiGCN~\cite{tong2020digcn} to $\mathcal{O}(nk)$ by simply adjusting the teleport probability $\alpha$ without any modification of directed graph structure $\tilde{\textbf{P}}$. They iteratively employ power method by rewriting the computation formula of $\pi_{appr}$ from DiGCN to the form of discrete-time Markiv chain: $\pi_{appr}^{t+1}=(1-\alpha)\pi_{appr}^t\tilde{\textbf{P}}+\frac{\alpha}{n}\pi_{appr}^t\textbf{1}^{n\times n}$. Thus, they only need to perturb the teleport probability $\alpha$ without any change of directed graph structure $\tilde{\textbf{P}}$ to generate another view for the graph contrastive learning.

\noindent
Recent advancements in directed GNNs have sought to enhance graph filter bases and eigenpolynomials by incorporating spectral information within the complex domain. This approach leverages the expressive power of complex representations to encode richer spectral characteristics, offering an alternative perspective to traditional methods grounded in the real domain. By utilizing complex-valued eigenstructures, these models aim to capture more nuanced relationships within graph data, potentially leading to improved model performance and theoretical representation capacity.
MagNet~\cite{zhang2021magnet} integrates the directionality of edges into the eigenpolynomial-based framework by designing a complex graph filter basis, also known as the magnetic Laplacian, where the undirected graph topology is encoded in the magnitude of entries and the directionality is encoded in the phase. Specifically, they formulate the graph filter basis as $\textbf{H}^{(q)}=\textbf{A}_s \odot exp(i\mathbf{\Theta}^{(q)})$, where $\textbf{A}_s=\frac{1}{2}(\textbf{A}(u,v)+\textbf{A}(v,u))$ and $\mathbf{\Theta}^{(q)}(u,v)=2\pi q(\textbf{A}(u,v)-\textbf{A}(v,u))$. The hyper-parameter $q$ determines the direction-encoding ability of magnetic Laplacian. For example, when $q=0$ and $\textbf{H}^{(0)}=\textbf{A}_s$, it means that the graph filter basis completely discards the directionality information; when $q=0.25$ and $\textbf{H}^{(0.25)}(u,v)=\frac{i}{2}=-\textbf{H}^{(0.25)}(v,u)$, it means that the directionality information is encoded in the phase of the graph filter basis.
MGC~\cite{zhang2021mgc} achieves the directed graph representation learning by enhancing the eigenpolynomial. Similar to MagNet~\cite{zhang2021magnet}, they adopt magnetic Laplacian operator as the graph filter basis: $\textbf{P}=\tilde{\textbf{D}}_s^{-\frac{1}{2}}\tilde{\textbf{A}}_s\tilde{\textbf{D}}_s^{-\frac{1}{2}}\odot \textbf{T}_q, \tilde{\textbf{D}}_s=\sum_{(u,v)\in \mathcal{E}}\tilde{\textbf{A}}_s(u,v), \tilde{\textbf{A}}_s=\textbf{A}_s+\textbf{I}, \textbf{T}_q(u,v)=exp(i2\pi q(\textbf{A}(u,v)-\textbf{A}(v,u)))$. In order to obtain a low-pass filter for the homogeneous graphs and a high-pass filter for the heterogeneous graphs, instead of exploiting Chebyshev polynomial, they introduce LinearRank as the eigenpolynomial: $\textbf{T}_{LR}=\frac{2}{K(K+1)}(K\textbf{I}-(K+1)\textbf{P}+\textbf{P}^{k+1})(\textbf{I}-\textbf{P})^{-2}$.
Framelet-MagNet~\cite{lin2023framelet_gnn} improves the graph filter basis to realize directed graph representation learning by applying a framelet transform to the directed graph signals to form a more sophisticated representation for filtering. Firstly, they adopt the magnetic Laplacian operator of MagNet~\cite{zhang2021magnet} to transform graph information from the spatial domain into the magnetic spectral domain. Then, they propose the Magnetic Graph Framelet Transform (MGFT) to transform magnetic graph signals to the framelet frequency domain for filtering, formulated as: $\rho_{n,s}^{(q)}(m) = \sum_{k=0}^{N-1}\textbf{u}_k(m)\zeta_0(\frac{\lambda_k}{2^s})\textbf{u}_k^*(n), \varrho_{n,s,r}^{(q)}(m) = \sum_{k=0}^{N-1}\textbf{u}_k(m)\zeta_r(\frac{\lambda_k}{2^s})\textbf{u}_k^*(n)$,
where $1\le r\le R$ is the framelet indexes, $\rho_{n,s}^{(q)}(m)$ and $\varrho_{n,s,r}^{(q)}(m)$ are the magnetic graph framelets for low-pass filter and high-pass filter respectively, $\left\{ (\textbf{u}_k, \lambda_k) \right\}_{k=0}^{N-1}$ are the eigenvector and eigenvalue pair of magnetic Laplacian operator $\textbf{L}^{(q)}$, $Z=\left\{ \zeta_0,\cdots ,\zeta_R \right\}$ and dilation level $s=1,\cdots ,S$ are hyper-parameters for differentiating frequencies into corresponding components.

\noindent
Differing from the real-value and the comlex-value directions mentioned above, another direction seek to introduce the holomorphic functional calculus into the design of the graph filter bases and eigenpolynomials to solve the obstacles in the graph Fourier transform that the graph filter bases without the real-symmetric property cannot be decomposed into a complete orthogonal eigenvectors, promoting the eigenpolynomial-based GNNs to a higher level.
HoloNet~\cite{koke2023holonets} enhances the graph filter basis and the eigenpolynomial altogether to achieve representation learning on directed graphs. They propose two sets of filters $\left\{ g_{\theta}^{fwd}(\textbf{T}) \right\}$ and $\left\{ g_{\theta}^{bwd}(\textbf{T}^*) \right\}$ based on the characteristic operator $\textbf{T}$ and its adjoint $\textbf{T}^*$ to allow information flow into two directions along the edges (e.g. forward and backward directions). More concretely, these forward- and backward-filters can be formulated as $g_{\theta}^{fwd/bwd}(\lambda)=\sum_{i\in I^{fwd/bwd}}\theta_i^{fwd/bwd}\mathbf{\Psi}_i^{fwd/bwd}(\lambda)$, where $\left\{ \mathbf{\Psi}_i^{fwd/bwd} \right\}_{i\in I^{fwd/bwd}}$ are different filter bases selected from the filter banks targeting at different tasks. For example, in the single graph scenario (e.g. node classification), the filter banks can be selected as: $\mathbf{\Psi}_i(\lambda_i)=\lambda^i$; while in the multiple graphs scenario (e.g. graph classification), the filter banks can be selected as: $\mathbf{\Psi}_i(\lambda_i)=\theta_i(\lambda^i-y)^{-i}$ where $y$ is a hyper-parameter. Moreover, HoloNet choose the Faber polynomial~\cite{ellacott1983computation} as the eigenpolynomial, which generalizes the traditional Chebyshev polynomial to subsets $U$ of the complex plane. 

\subsection{Sequence-based Framework}
\label{sec:sbf}
\textbf{Motivations.} The sequence-based framework in GNNs draws inspiration from Transformer, reinterpreting structured graphs as sequences of tokens (e.g. nodes and subgraphs). Initially, tokenization is applied to represent the graph as a series of elements. For instance, a node tokenizer, denoted as $\Upsilon(G)=\left\{ \textbf{P}_{v_1}, \textbf{P}_{v_2}, \cdots ,\textbf{P}_{v_{|\mathcal{V}|}} \right\}$, tokenizes each node $v_i$ as a numerical token $\textbf{P}_{v_i}$. Alternatively, the subgraph tokenizer applies a similar approach represented by $\Phi(G)=\left\{ \textbf{P}_{s_1}, \textbf{P}_{s_2}, \cdots ,\textbf{P}_{s_T} \right\}$. Following the tokenization step, the encoding step would be applied to capture the long-range dependencies among item pairs in the sequence: $\textbf{H}^{(i)}=\Psi_i(AGG_i(\Upsilon(G) / \Phi(G)))$, where $\Psi_i$ and $AGG_i$ are the $i$-th encoder and the $i$-th aggregator, respectively. The sequence-based framework aims to refine the tokenizer and encoder mechanisms and to strengthen the model's capability to  learn from such graph representations.

\noindent
\textbf{Methods.}
PACE~\cite{dong2022pace} enhances the node tokenizer within the sequence-based framework by introducing the dag2seq framework, which efficiently transforms DAGs into unambiguous sequences of node tokens. This approach facilitates expressive directed graph representation learning by effectively capturing the unique, long-range dependencies inherent in DAGs. Unlike traditional methods that utilize topological order as the sequence ordering, PACE adopts a canonical order in conjunction with a Transformer encoder and a one-layer injective GNN. This setup ensures injectivity between the original DAGs and the generated node sequences, preserving structural fidelity and supporting robust downstream learning tasks.
TMDG~\cite{geisler2023transformers_meet_digraph} introduces two encoding techniques within the sequence-based framework using the node tokenizer and subgraph tokenizer to enable models to effectively capture long-range relationships among node and subgraph tokens while maintaining sensitivity to directed structural topology. The first encoding technique incorporates the Magnetic Laplacian, following the Magnet method~\cite{zhang2021magnet}, by selecting an appropriate hyper-parameter $q$ for positional encoding, thereby enabling the model to reflect graph directionality. The second technique extends the concept of traditional random walks on undirected graphs by integrating bidirected random walks (i.e., both forward and backward directions) into the positional encoding of the sequence-based framework. These encoding mechanisms aim to enrich the model's ability to recognize directed dependencies and structural nuances within graph representations. 
DAGRA~\cite{luo2024dagformer} also enhances the encoding mechanism of the sequence-based framework using the node tokenizer to allow the model to capture the long-range directed dependencies in the Directed Acyclic Graphs (DAGs). Observing the fact that a node's predecessors and successors are more relative to itself than other nodes, they restrict the reception field of each node's attention mechanism to: $N_k(v_i)=\left\{ (v_i,v_j)\in \le_k \right\}\cup \left\{ (v_j,v_i)\in \le_k \right\}$ where $\le_k$ means paths that are less than $k$-hop away to node $v_i$.
ADR-GNN~\cite{eliasof2023adr} improves the node tokenizer of the sequence-based framework by introducing the Advection-Diffusion-Reaction systems into directed GNNs. They discretize the formulas of the Advection-Diffusion-Reaction systems in the physical world and propose the corresponding principles on directed graphs. Among them, the advection operator, the diffusion operator and the reaction operator are responsible for transporting node features along the directed edges, aggregating node features from different directed edges and pointwise reactions, respectively.
DiGT~\cite{wangdirected} upgrades the encoding mechanism within the sequence-based framework by employing a node tokenizer that incorporates dual encodings~\cite{kollias2022nste} for each node. This approach extends the foundational concepts of the Graph Transformer~\cite{cai2020graph} by introducing distinct $\textbf{Q},\textbf{K},\textbf{V}$ matices for source and target nodes individually. Through this separation, attention matrices are computed independently for source and target nodes, enabling a more nuanced encoding of directed relationships. The final node embeddings, derived from these attention matrices, are subsequently utilized in downstream tasks, effectively capturing complex interactions across nodes with directed specificity.
MMLPE~\cite{huang2024good} enhances the encoding mechanism of node-tokenizer-utilized sequence-based framework by proposing Multi-$q$ Magnetic Laplacian Postional Encoding (Multi-$q$ Mag-PE). Firstly, they propose the notion of walk profile and illustrate the existing PE mechanisms fail to distinguish different walk profiles on the directed graphs, including Mag-PE~\cite{zhang2021magnet} and SVD-PE~\cite{hussain2022global}. Then, they propose the Multi-$q$ Mag-PE by leveraging different $q$ simultaneously based on the frequency interpretation that different $q$ act like different frequencies recording the accumulated phase shift.
DiRW~\cite{su2024dirw} adopts an alternative approach to enhance the subgraph tokenizer by integrating directed random walks into the sequence-based framework. Initially, a direction-aware path sampler is implemented, which considers walk probabilities, the number of walks, and walk lengths in a weight-free manner. Subsequently, node embeddings are generated by adaptively aggregating the various sampled paths through a learnable process. This methodology facilitates a more nuanced representation of the graph structure, and a more effective capacity of capturing directed relationships within the graph data.

\section{Revisiting Directed GNNs from the View of Data-centric Machine Learning}
\label{sec:revisit}
As outlined in Section~\ref{sec:intro}, extensive former researches have primarily focused on \textbf{model-centric} approaches in graph learning, emphasizing model architecture development while often overlooking the critical role of the graph data itself. However, substantial evidence suggests that both the nuanced understanding of graph structure and the quality of graph data can significantly impact model performance. Consequently, this section adopts a \textbf{data-centric} perspective to re-evaluate existing directed GNNs, investigating how these models interpret and refine graph data to enhance their effectiveness and predictive accuracy, as shown in Table~\ref{tab:dcv}. First, we begin by introducing three primary views for examining graph data, which form the conceptual foundation for designing and refining directed GNNs. Following the discussion of these graph views, we explore various techniques for graph data improvement, focusing on topological enhancement and node feature enhancement. These enhancement techniques, tailored to different graph views, provide critical strategies for increasing the expressiveness and performance of directed GNN models.

    \begin{table*}[t]
    \centering
    \caption{Revisited Directed GNNs from the Data-centric View.}
    \label{tab:dcv}
    \resizebox{\textwidth}{!}{
    \begin{tabular}{@{}cccccccccc@{}}
    \toprule
    \multirow{3.4}{*}{Models} & \multirow{3.4}{*}{Venue} & \multirow{3.4}{*}{\makecell[c]{Directed Graph \\Data Understanding}} & \multicolumn{7}{c}{Directed Graph Data Improvement}                                                                                                                                                  \\ \cmidrule{4-10}
                           &                        &                                     & \makecell[c]{Multigraph?} & \makecell[c]{Acyclic\\ Graphs?} & \makecell[c]{Random\\ Graphs?} & \makecell[c]{Augmented\\ Graphs?} & \makecell[c]{Diffused\\ Graphs?} & \makecell[c]{Augmented\\ features?} & \makecell[c]{Denoised\\ features?} \\ \midrule
          MotifNet~\cite{monti2018motifnet}          &    DSW'18                    &      Graph Spectral View                               &                      &                        &                        &      \checkmark                      &                           &                          &                         \\
          DGCN~\cite{ma2019spectral}                 &      arXiv'19                  &       Graph Spectral View                  &                      &                         &                        &        \checkmark                    &                           &                          &                         \\
          D-VAE~\cite{zhang2019dvae}                 &       NeurlPS'19                 &       Graph Topological View              &                      &        \checkmark                 &                        &                            &                           &                          &                         \\
         edGNN~\cite{jaume2019edgnn}                  &       ICLR'19                 &         Graph Topological View             &               &                         &                        &                            &                           &                          &                         \\
          FDGCN~\cite{li2020scalable}                 &       Access'20                 &        Graph Spectral View             &                      &                         &      \checkmark                  &                            &                           &                          &                         \\
          DiGCN~\cite{tong2020digcn}                 &        NeurlPS'20                &       Graph Spectral View             &                      &                         &                        &        \checkmark                    &                           &                          &                         \\
          DGCN~\cite{tong2020dgcn}                 &     arXiv'20                   &        Graph Topological View                &                      &                         &                        &         \checkmark                   &                           &                          &                         \\
          MagNet~\cite{zhang2021magnet}                 &      NeurlPS'21                  &      Graph Spectral View               &                      &                         &      \checkmark                  &       \checkmark                     &                           &                          &                         \\
          DiGCL~\cite{tong2021directed}                 &       NeurlPS'21                 &       Graph Spectral View            &                      &                         &                        &         \checkmark                   &                           &                          &                         \\
          DAGNN~\cite{thost2021dagnn}                 &       ICLR'21                 &        Graph Topological View              &                      &         \checkmark                &                        &                            &                           &                          &                         \\
          MGC~\cite{zhang2021mgc}                 &     arXiv'22                   &        Graph Spectral View             &                      &                         &                        &                           &                           &                          &       \checkmark                  \\
          NSTE~\cite{kollias2022nste}                 &     AAAI'22                   &        Graph Topological View            &                      &                         &                        &                            &                           &        \checkmark                  &                         \\
          GRANDE~\cite{wu2022grande_directed_app_financial2}                 &     ICDM'22                   &      Graph Topological View            &       \checkmark               &                         &                        &                            &                           &                          &                         \\
          D-HYPR~\cite{zhou2022dhypr}                 &     CIKM'22                   &      Graph Topological View             &                      &                         &                        &      \checkmark                      &                           &        \checkmark                  &                         \\
         DIMPA~\cite{he2022dimpa}                  &     LoG'22                   &        Graph Topological View               &                      &                         &        \checkmark                &       \checkmark                     &                           &         \checkmark                 &                         \\
        PACE~\cite{dong2022pace}                   &     ICML'22                   &       Graph Sequence View              &                      &          \checkmark               &                        &                            &       \checkmark                    &                          &                         \\
        Framelet-MagNet~\cite{lin2023framelet_gnn}                   &     ICASSP'23                   &        Graph Spectral View     &                      &                         &                        &       \checkmark                     &                           &                          &                         \\
        HoloNet~\cite{koke2023holonets}                   &      arXiv'23                  &        Graph Spectral View              &                      &                         &                        &      \checkmark                      &                           &                          &                         \\
         Dir-GNN~\cite{dirgnn_rossi_2023}                  &    LoG'23                    &     Graph Topological View       &                      &                         &      \checkmark                  &                            &                           &                          &                         \\
        A2DUG~\cite{maekawa2023a2dug}                   &      arXiv'23                  &       Graph Topological View      &                      &                         &                        &      \checkmark                      &                           &                          &                         \\
        TMDG~\cite{geisler2023transformers_meet_digraph}                   &     ICLR'23                   &        Graph Sequence View          &                      &                         &                        &                            &      \checkmark                     &                          &                         \\
         DAGRA~\cite{luo2024dagformer}                   &      NeurlPS'23                  &      Graph Sequence View           &                      &    \checkmark                     &                        &                            &           \checkmark                &                          &                         \\
         ADR-GNN~\cite{eliasof2023adr}                   &       arXiv'23                 &      Graph Sequence View      &                      &                         &                        &                            &       \checkmark                    &                          &                         \\
        LightDiC~\cite{li2024lightdic}                   &     VLDB'24                   &      Graph Topological View      &                      &                         &                        &          \checkmark                  &                           &                          &                         \\
        ADPA~\cite{sun2023adpa}                   &       ICDE'24                 &        Graph Topological View      &                      &                         &                        &         \checkmark                   &                           &                          &                         \\
         GNNDM~\cite{egressy2024provably_directed_app_financial3}                  &       AAAI'24                 &       Graph Topological View           &        \checkmark  &                         &                        &                            &                           &                          &                         \\
         DiGT~\cite{wangdirected}                  &      TMLR'24                  &       Graph Sequence View        &                      &                         &     \checkmark                   &                            &     \checkmark                      &       \checkmark                   &                         \\
        MMLPE~\cite{huang2024good}                   &    arXiv'24                    &      Graph Sequence View         &                      &                         &                        &         \checkmark                   &     \checkmark                      &                          &                         \\
        DiRW~\cite{su2024dirw}                   &     arXiv'24                   &       Graph Sequence View     &                      &                         &                        &                            &     \checkmark                      &                          &                         \\ \bottomrule
    \end{tabular}}
    \end{table*}

\subsection{Graph Data Understanding \& Exploitation}
\label{sec:gdue}
The initial step in designing a GNN is to develop a robust understanding of the underlying graph data. Graph data can be examined from several perspectives, such as topological, spectral, and sequential views, each providing a distinct interpretation and offering unique insights into the data’s structure. These varied perspectives influence the architecture of directed GNNs in different ways. Traditional model-centric approaches often treat these views interchangeably, which may overlook the distinct informational contributions each perspective provides. For example, models might be primarily informed by topological insights yet reinterpreted through a spectral lens, potentially diluting the strengths of each view. In contrast, a data-centric approach to graph machine learning recognizes that these different perspectives are integral to informing innovative model design. Each view—whether topological, spectral, or sequential—reveals unique structural or relational aspects of the graph data, which can guide the development of more expressive and effective directed GNN models. This section will elaborate on the intrinsic characteristics of these perspectives, aiming to advance the graph learning field by fostering models that better capture the nuanced information within graph data.

\noindent
\textbf{Graph Topological View.} Viewing a graph as a composition of nodes and edges represents the most intuitive and direct approach to understanding its structure. Unlike Euclidean data, which possesses a regular topology and fixed dimensionality, graph data is irregular and cannot directly leverage traditional neural network architectures, such as Convolutional Neural Networks (CNNs) commonly used in the domain of computer vision. This structural distinction has led to the development of the message-passing framework, specifically designed to facilitate machine learning on irregular, non-Euclidean data. Numerous directed GNNs~\cite{jaume2019edgnn, zhang2019dvae, tong2020dgcn, thost2021dagnn, kollias2022nste, li2024lightdic, zhou2022dhypr, he2022dimpa, maekawa2023a2dug, dirgnn_rossi_2023, sun2023adpa, wu2022grande_directed_app_financial2, egressy2024provably_directed_app_financial3} utilize message-passing framework to learn directed graph representations, as outlined in Section~\ref{sec:mpf}. These models conceptualize the graph as a set of nodes connected by directed edges and introduce various types of propagation/aggregation mechanisms within message-passing framework to enable efficient information exchange across graph.

\noindent
\textbf{Graph Spectral View.} Viewing a graph through the lens of a discretized graph spectrum provides a relatively abstract approach to understanding its intrinsic properties. This perspective is grounded in graph spectral theory, which employs the Graph Fourier Transform to convert spatial graph topology into a combination of orthogonal graph spectra (i.e., eigenvectors of the graph filter basis). Given its theoretical rigor and strong empirical performance, many directed GNNs~\cite{monti2018motifnet, ma2019spectral, li2020scalable, tong2020digcn, zhang2021magnet, tong2021directed, zhang2021mgc, lin2023framelet_gnn, koke2023holonets} adopt this graph spectral view to capture the essence of graph data, as illustrated in Section~\ref{sec:ebf}. Unlike conventional approaches in signal processing, these models design diverse graph filter bases and eigenpolynomials tailored to specific graph spectra, guided by the principles of the Graph Fourier Transform. Moreover, inspired by the Graph Fourier Transform’s capacity to project data across domains, these directed GNNs extend typical transformations by projecting graph data from the real-value domain to complex-value domains (e.g., MagNet~\cite{zhang2021magnet} and MGC~\cite{zhang2021mgc}) and from the magnetic spectrum domain to the framelet spectrum domain (e.g., Framelet-MagNet~\cite{lin2023framelet_gnn}). This advanced approach enriches the graph's representational capacity, providing additional new avenues for modeling more complex directed graph structures.

\noindent
\textbf{Graph Sequence View.} Viewing a graph as a sequence is an innovative perspective inspired by Transformer. Unlike the topological and spectral views, graph sequence view de-emphasizes explicit topology by representing graphs as sequences of tokens, such as nodes or subgraphs. This transformation converts the irregular, non-Euclidean graph data back to a regular Euclidean format, facilitating the capture of long-range dependencies between token pairs. However, graph sequence view does not disregard graph topology entirely. Instead, it employs positional encoding to integrate structural information when applying aggregation mechanisms. Recent directed GNNs have increasingly adopted this view, including models such as PACE~\cite{dong2022pace}, TMDG~\cite{geisler2023transformers_meet_digraph}, DAGRA~\cite{luo2024dagformer}, DiGT~\cite{wangdirected}, and MMLPE~\cite{huang2024good}, as illustrated in Section~\ref{sec:sbf}. These models interpret graph data as a sequence of node tokens and leverage various attention mechanisms to capture relationships across nodes separated by multiple hops. Differing from these approaches, DiRW~\cite{su2024dirw} represents graph data as a sequence of subgraph tokens, using random walks as the token units. Attention mechanisms are then applied to model relationships between these subgraph tokens, providing a nuanced view of long-range subgraph dependencies. These advanced approaches significantly expand the ability to capture long-range dependencies within directed graphs, opening up new possibilities for advancing GNN architectures and their applications.

\subsection{Graph Data Improvement}
\label{sec:GDI}
Unlike image or text data, where data points are embedded within regular grid structures, graph-structured data represent relationships among data points through irregular topologies. To address challenges arising from low-quality graph structures, the research community has developed and investigated various data enhancement techniques that draw inspiration from different graph views illustrated in Section~\ref{sec:gdue} and are aimed at refining and improving the underlying topology and node embeddings in graph representations. In this subsection, we formulate and review two directions of graph data improvement technologies: topological enhancement and node feature enhancement.

\noindent
\subsubsection{Topological Enhancement}
To enhance the topological structure of graph data, substantial efforts have been directed towards this area of research. As outlined in the data-centric graph learning pipeline illustrated in Fig.~\ref{fig: data-centric}, the stages of graph construction and pre-processing present valuable opportunities for improvement. For instance, the original graph data can be transformed into alternative graph structures, such as multigraphs, directed acyclic graphs, or random graphs, to better capture node relationships specific to an application domain. Additionally, various topological modification techniques can be applied to refine directed graphs, including structural augmentation methods and diffusion-based approaches, to optimize the graph representation and improve model performance.

\noindent
\textbf{Methods for Directed Multigraphs.}
In numerous real-world applications, such as financial transactions, social networks, and biological systems, multiple interactions between the same pair of nodes are common. This prevalence has underscored the growing significance of directed multigraphs, which are an extension of traditional directed graphs that permit multiplicity in edges.
For instance, GRANDE~\cite{wu2022grande_directed_app_financial2} and GNNDM ~\cite{egressy2024provably_directed_app_financial3} adopt directed multigraph modeling to represent financial transaction networks more accurately, as this approach captures the multiplicity of relationships (i.e., multiple transactions) between specific nodes. Moreover, these models also design a tailored message-passing framework that enables the transfer of information across the multiple edges between two nodes, thus enhancing their ability to reflect the complex relational structures within the data.

\noindent
\textbf{Methods for Directed Acyclic Graphs.}
Directed acyclic graphs are a fundamental data structure prevalent in various domains, including source code, logical formulas, and architecture of a neural network.
They represent computations and processes that have a clear starting point and a well-defined sequence of operations without cycles. With the rise of GNNs, there is a growing interest in extending their capabilities to handle DAGs effectively.
The motivation behind developing specialized GNNs for DAGs stems from their unique partial orders defined by their edges, which represents the sequence of operations or dependencies.
Traditional model-centric directed GNNs do not inherently respect this order and may not capture the full computational semantics of DAGs.
This has led to the development of D-VAE~\cite{zhang2019dvae}, DAGNN~\cite{thost2021dagnn}, PACE~\cite{dong2022pace} and DAGRA~\cite{luo2024dagformer}, which are designed to leverage the structural and computational properties of DAGs for improved learning and reasoning.
The former two directed GNNs develop corresponding message-passing scheme that is tailored to the scenarios of DAGs, while the latter two introduce sequence-based framework into the DAGs machine learning. They are all taking partial order, the special property of DAGs, into account, thus enhancing their ability to capture intrinsic relationship within the structures of DAGs.

\noindent
\textbf{Methods for Directed Random Graphs.} To enhance the performance and robustness of directed GNNs during training, particularly given the scarcity and limited quality of graph data, the application of random graphs as a topological enhancement technique has become essential. Several directed GNNs leverage directed random graphs to strengthen model expressiveness and stability in directed graph learning. For instance, FDGCN~\cite{li2020scalable} utilizes the Mixed Membership Stochastic Block Model (MMSBM) to generate random graphs, which are incorporated during additional training epochs to enrich the training data. Similarly, MagNet~\cite{zhang2021magnet}, DIMPA~\cite{he2022dimpa}, and DiGT~\cite{wangdirected} apply the Directed Stochastic Block Model (DSBM). This model takes edge existence and directionality into account, governed by two hyperparameters, $\alpha$ and $\beta$, to generate directed graphs that align with the model’s training requirements. Additionally, Dir-GNN~\cite{dirgnn_rossi_2023} generates random graphs with varied levels of homophily by employing a modified preferential attachment process, supporting diverse structural properties in the directed graph data.

\noindent
\textbf{Methods for Directed Augmented Graphs.} After the graph is constructed from the original data, various kinds of graph augmentation can be further applied in the pre-training process to obtain the more appropriate graph topology for the following training process. MotifNet~\cite{monti2018motifnet} alters the original graph topology by introducing the induced motif matrix to facilitate the graph machine learning on the directed graphs. Inspired by the station distribution of random walk, DGCN~\cite{ma2019spectral} reweights the original graph adjacency matrix by $\textbf{P}=\textbf{D}_{out}^{-1}\textbf{A}$ to ease the over-smoothing problem and achieve directed graph machine learning. Differing from the reweighting technique mentioned above, DGCN~\cite{tong2020dgcn}, DIMPA~\cite{he2022dimpa} and D-HYPR~\cite{zhou2022dhypr} propose and encode the $k$-order proximity into the adjacency matrix, to incorporate features from neighbors of multiple hops and extend the reception field of the model. Drawing the inspiration from personalized PageRank, DiGCN~\cite{tong2020digcn} and DiGCL~\cite{tong2021directed} augment the original graph topology by employing the approximate directed Laplacian operator to facilitate more powerful and expressive directed graph learning while lower the computation cost. MagNet~\cite{zhang2021magnet}, Framelet-Magnet~\cite{lin2023_framelet-magnet}, HoloNet~\cite{koke2023holonets}, LightDiC~\cite{li2024lightdic} and MMLPE~\cite{huang2024good} enhance the original graph adjacency matrix through replacing the real-value in the matrix by complex-value and encoding the edge directionality into the phase of the complex element. A2DUG~\cite{maekawa2023a2dug} and ADPA~\cite{sun2023adpa} adopt a similar graph augmentation technique by introducing multi-hop propagator to alter the original graph adjacency matrix. While A2DUG employ shallow propagators which ignore the directionality of edges, ADPA proposes a more advanced propagators which encode edge directionality into the various Directed Patterns (DPs).

\noindent
\textbf{Methods for Directed Diffused Graphs.} Diffusion-based structural enhancement techniques take inspiration from the graph sequence view illustrated in Section~\ref{sec:gdue}, which conceptualizes a graph as a sequence of node or subgraph tokens. By employing various augmentation mechanisms, these techniques aim to capture long-range dependencies among tokens. Recently, directed GNNs have increasingly focused on diffusion-based structural enhancement. Approaches such as PACE~\cite{dong2022pace}, TMDG~\cite{geisler2023transformers_meet_digraph}, ADR-GNN~\cite{eliasof2023adr}, DiGT~\cite{wangdirected}, DAGRA~\cite{luo2024dagformer}, and MMLPE~\cite{huang2024good} augment the original graph topology by implementing node-token diffusion mechanisms. Similar to the Transformer architecture, they utilize various attention mechanisms to capture long-range relationships between node pairs, adapting the adjacency matrix into an attention matrix for subsequent training.
An alternative approach is taken by DiRW~\cite{su2024dirw}, which uses diffusion-based enhancement by viewing graphs as sequences of path subgraphs sampled from random walks. In this approach, the original adjacency matrix is transformed into an augmented matrix, derived from aggregated embeddings of diverse random walks, to better capture structural nuances across paths.

\subsubsection{Node Feature Enhancement}
In graph-structured data, node features serve as critical elements that convey the semantic information of each component. For example, in citation networks, node features are often represented as word-bag vectors that capture the content of each paper, providing valuable semantic insights for downstream tasks. However, the quality of these features in real-world datasets is often suboptimal. Also in citation networks, for instance, extracted features may lack significant terms or include extraneous ones, leading to issues of missing or noisy features that can adversely impact the performance of graph-based learning models. To enhance the quality of node features in directed graph data, two kinds of primary data-centric approaches have been actively pursued in recent years: node feature augmentation and node feature denoising. The focus of node feature augmentation lies in supplementing or reconstructing appropriate node features that align well with the training objectives of directed GNNs. In contrast, node feature denoising is primarily concerned with removing extraneous or misleading features that could adversely affect the training performance of directed GNNs. Together, these methods aim to improve feature reliability, facilitating more accurate and robust model performance through the modification of node features.

\noindent
\textbf{Methods for Directed Augmented Node features.} Node feature augmentation primarily aims to enhance or reconstruct node features to better align with the specific training objectives of directed GNNs. This process seeks to enrich node representations by either supplementing missing feature data or modifying the existing feature vectors that capture relevant information, ultimately supporting improved model performance in directed graph tasks. Recent research has increasingly emphasized the development of more expressive and robust directed GNNs through advanced node feature augmentation techniques. For instance, DIMPA~\cite{he2022dimpa}, NSTE~\cite{kollias2022nste} and DiGT~\cite{wangdirected} enhance node features by introducing dual encoding, which splits the original node features into distinct components (e.g., source and target) to generate enriched features tailored to directed graph contexts. D-HYPR~\cite{zhou2022dhypr} further advances node feature augmentation by mapping original features from Euclidean space into hyperbolic space, leveraging the complex, non-Euclidean latent anatomy provided by hyperbolic projection to support more effective directed graph learning.

\noindent
\textbf{Methods for Directed Denoised Node features.} Node feature denoising aims to remove erroneous or potentially detrimental features that could impair the training and performance of directed GNNs. This approach seeks to identify and filter out noisy data elements that may obscure essential patterns within the graph, thereby enhancing the model’s capacity to learn meaningful representations and bolstering robustness in directed graph learning tasks. Although research in this area remains limited, investigating node feature enhancement through denoising holds significant promise for improving model reliability and interpretability in directed GNNs. Drawing inspiration from the graph spectral view illustrated in Section~\ref{sec:gdue}, MGC~\cite{zhang2021mgc} denoises the node features by minimizing the graph signal denoising formula as $\min_{\textbf{x}}\left\{ \mu \Vert\Bar{\textbf{x}}-\textbf{x}\Vert_2^2+\mathcal{S}_p(\textbf{x}) \right\}$, where $\textbf{x}=\Bar{\textbf{x}}+\textbf{n}$ is the noisy graph signal composed of the pure part $\Bar{\textbf{x}}$ and the noisy part $\textbf{n}$, $\mu >0$ is the trade-off coefficient and $\mathcal{S}_p(\textbf{x})$ is the Dirichlet energy~\cite{cai2020note}.

\section{Industrial Applications of Directed GNNs}
\label{sec:applications}
Directed graph representation learning is crucial in the scenarios of widespread industrial applications. Drawing on a thorough review of existing literature in graph machine learning, we classify these real-world applications by identifying the specific tasks associated with directed graph embedding, with a particular emphasis on nodes, directed edges, and the overall graph structure (e.g., node-level, edge-level and graph level, respectively). It is important to note that each study is associated solely with the tasks explicitly described in the respective articles.
\subsection{Node-level Applications}
In this section, we elucidate the role of directed graph representation learning in the context of industrial applications defined on the node-level of a directed graph, such as classification, regression, and recommendation. For each task, we additionally expound on the specific application domains in which it has been employed.

\noindent
\textbf{Node Classification.} The primary objective of the node classification task is to accurately assign labels to the unlabeled nodes within a graph by leveraging the label information provided for a subset of labeled nodes. This paper presents two representative applications: phishing detection within the Ethereum network (GAE\_PDNA~\cite{ratra2024graph_directed_app_pd2}, GrabPhisher~\cite{zhang2024grabphisher_directed_app_pd3}, SIEGE~\cite{li2023siege_directed_app_pd6},
BM-GCN~\cite{zhang2023phishing_directed_app_pd7}, TransWalk~\cite{xiong2023ethereum_directed_app_pd8}, GL4AML~\cite{karim2024scalable_directed_app_pd9}) and conversational emotion recognition (GNN4CER~\cite{gan2024graph_directed_app_er1}, DialogueGCN~\cite{ghosal2019dialoguegcn_directed_app_er2}, DAG-ERC~\cite{shen2021_directed_app_er3}, GraphCFC~\cite{li2023graphcfc_directed_app_er4}, COGMEN~\cite{joshi2022cogmen_directed_app_er5}, GA2MIF~\cite{li2023ga2mif_directed_app_er6}, ProtoDAG~\cite{kang2023directed_directed_app_er7}, CER-GNN~\cite{lian2020conversational_directed_app_er8}). In the context of phishing account detection within the Ethereum network, Ethereum is represented as a directed graph, where nodes correspond to individual accounts, edges signify transactions, and the direction of the edges reflects the flow of transactions. Graph neural networks classify the nodes as either normal accounts or phishing accounts by leveraging the embeddings derived from the nodes. To facilitate the recognition of emotions within individual utterances of a conversation, the entire conversational exchange is configured as a directed graph. In this graph-based model, each node corresponds to an individual utterance, and each directed edge delineates the directed flow of information between these utterances. Furthermore, the type of the edges are utilized to ascertain whether the connected utterances originate from the same speaker. Subsequently, the graph neural network infers the emotional labels for each utterance based on the node embeddings associated with them.

\noindent
\textbf{Node Regression.} The task of node regression is to predict a continuous numerical value associated with each node within a network. This process leverages node-specific features, structural information, or a combination of both to develop a predictive model. Our investigation reveals that node regression on directed graphs holds significant potential for applications in traffic prediction, particularly in traffic flows prediction (T-DGCN\cite{chen2021temporal_directed_app_tf1}, Traffic-GCNN\cite{wang2022traffic_directed_app_tf2}, STGNN\cite{li2022spatio_directed_app_tf3}, SeqGNN\cite{xie2019sequential_directed_app_tf4}, DGNNs\cite{sharma2023graph_directed_app_tf5}, STGC-GNNs~\cite{he2023stgc_directed_app_tf6}, AGP-GNN~\cite{guo2023adaptive_directed_app_tf7}, TF-MVGNN~\cite{cheng2024tf_directed_app_tf8}) and pedestrian trajectory forecasting (TFDGCNN\cite{su2022trajectory_directed_app_ptf1}, TDAGCN\cite{wang2023trajectory_directed_app_ptf2}, DBSTGNN\cite{zhang2023dual_directed_app_ptf3}, MDST-DGCN\cite{liu2022mdst_directed_app_ptf4}, SGCN\cite{shi2021sgcn_directed_app_ptf5}, MTP-GAN~\cite{mo2022multi_directed_app_ptf6}, HSTGA~\cite{alghodhaifi2023holistic_directed_app_ptf7}, XHGP~\cite{limeros2023towards_directed_app_ptf8}) . In traffic flow prediction, the problem is commonly formulated as a sequence of directed graphs, with each graph representing the traffic conditions at a specific time interval. Within these graphs, nodes correspond to individual road segments, while edges capture the connectivity between segments. The directionality of the edges signifies the flow of traffic across the network. Graph neural networks predict the probability of congestion for each segment by utilizing the node embedding representations associated with those segments. In contrast to traffic flow prediction, pedestrian trajectory prediction typically represents pedestrians as nodes and models the distances between pedestrians as edges. For each pedestrian, the graph neural network predicts the most probable position at a specified time by utilizing its embedding representation.

\noindent
\textbf{Node Recommendation.} The goal of node recommendation is to ascertain and endorse the top-k nodes that have strong relationships with a designated query node within a directed graph. For examples, in recommendation system, directed graphs serve as a potent instrument to structurally encode the complex, high-order relationships that extend beyond the traditional user-item interactions. Moreover, directed graphs are capable of explicitly modeling additional dimensions of user-item interactions, such as product asymmetry (DAEMON~\cite{virinchi2022_directed_app_recommendation2}, MERLIN ~\cite{tiady2024merlin_directed_app_recommendation3}, BLADE~\cite{Virinchi2023_directed_app_recommendation4}) . They can also represent supplementary interaction data, such as the co-occurrence between items (UGRec~\cite{zhao2021ugrec_directed_app_recommendation1}).

\subsection{Edge-level Applications}
In this section, we will explore the role of directed graph representation learning in relation to edge-level tasks such as link prediction and edge classification, as well as its relevant application domains.

\noindent
\textbf{Link Prediction.} Link prediction in directed graphs involves not only the needs of determining the presence of an interaction relationship between pairs of entities based on their attributes and currently observed links, but also the necessitates for forecasting the directed perspectives of these interactions. Our review indicates that the link prediction techniques utilizing directed graphs have broad applications in various application domains, such as the research of Gene Regulatory Networks, or GRNs for short (GRINCD~\cite{feng2023gene_directed_app_grn4}, GNNLink~\cite{mao2023predicting_directed_app_grn3}, GRN-IUGNN~\cite{jereesh2024reconstruction_directed_app_grn2}, GENELink
~\cite{chen2022graph_directed_app_grn5}). In GRNs, nodes typically correspond to genes or transcription factors, while directed edges represent the regulatory interactions and dependencies between these entities. GNNs facilitate the analysis of such regulatory relationships by predicting the existence and directionality of edges between nodes, thereby capturing the underlying regulatory dynamics among genes or transcription factors. This approach enables the identification and characterization of complex regulatory pathways within GRNs, contributing to a deeper understanding of biological systems.

\noindent
\textbf{Edge Classification.} The goal of the edge classification task is to accurately assign labels to the unlabeled edges by leveraging the information derived from a subset of labeled edges, in conjunction with the graph's topological structure. For examples, in the domain of financial network analysis, edge classification tasks utilizing directed graphs are applied to analyze the suspicious transactions within the financial networks(GRANDE~\cite{wu2022grande_directed_app_financial2}). More specifically, users are often regarded as nodes and transactions are often regarded as edges. The direction of an edge is typically understood as the direction of its corresponding cash flow. Graph neural networks categorize the edges within the transaction network as suspicious or normal transactions by analyzing the embedded information derived from the processed nodes and edges.

\subsection{Graph-level Applications}
In the ensuing discourse, we will delve into the application of directed graph representation learning in the graph-level tasks, including graph classification, graph regression, and graph matching. Furthermore, we will conduct a detailed examination of the industrial application scenarios pertinent to each task type.

\noindent
\textbf{Graph Classification.} The objective of the graph classification task is to assign labels to entire graphs by analyzing both their structural and feature information. In our investigation, numerous studies focus on the application domains of malware detection (SPNGNN~\cite{zapzalka2024semantics_directed_app_md1}, GNN-MDF~\cite{esmaeili2023gnn_directed_app_md2}, NF-GNN~\cite{busch2021nf_directed_app_md3}, NT-GNN~\cite{liu2023nt_directed_app_md4}, SDGNet~\cite{9200767liu2023nt_directed_app_md5}, MDGraph~\cite{li2024mdgraph_app_md6}, DawnGNNz~\cite{feng2024dawngnn_app_md7}) , social network analysis (Bi-GCN~\cite{bian2020_directed_app_social1}, OID-GCN~\cite{song2021rumor_directed_app_social3},  EBGCN~\cite{wei2021towards_directed_app_social4}, RoSGAS~\cite{yang2023rosgas_directed_app_social5}) , and brain network analysis (DBNA ~\cite{alaei2023_directed_app_bna1}, CGCNN~\cite{kong2022causal_directed_app_bna2}, Hodge-GNN~\cite{park2023convolving_directed_app_bna3}, TBDS ~\cite{yu2022learning_directed_app_bna4}, DSL-GNN\cite{cao2024dementia_directed_app_bna5}, DABNet~\cite{yu2023deep_directed_app_bna6}, TSGBF~\cite{wang2023eeg_directed_app_bna7}, SSGNN4ESA~\cite{tang2021self_directed_app_bna8}). In the domain of malware detection, directed graphs are commonly constructed from control flow graphs, where each node represents a basic block of instructions, and the edges signify interactions between these blocks. By employing graph neural networks for representation learning of the entire graph, malware detection can be conceptualized as a binary classification task applied to the graph. In the field of social network analysis, potential rumors are represented as graphs, wherein nodes correspond to individual posts associated with the rumor, and edges signify interaction relationships between these posts, such as responses or retweets. By leveraging graph neural networks, the entire rumor graph is analyzed to assess the veracity of the information. In brain network analysis, the nodes and edges of the directed graph are constructed based on the power spectrum density and effective brain connectivity, which are extracted from EEG data. Subsequent processing by graph neural networks enables the emotion recognition and disease detection based on EEG data to be conceptualized as graph classification problems.

\noindent
\textbf{Graph Regression.} Graph regression tasks involve the prediction of continuous numerical targets that are associated with the entirety of a graph. This is achieved by leveraging the node features, edge features, and the overarching topological structure of the graph. Our investigation reveals that the applications of graph regression techniques significantly enhance the performances of models in the area of chip design (GCPRMM\cite{ji2022gnn_directed_app_chip1}, GACPRMM~\cite{ji2023gat_directed_app_chip2}, GCPVIFR\cite{ji2024gnn_directed_app_chip4}) . The construction of a directed graph is derived from a given microfluidic mixer, the internal nodes and channels in the mixer, are systematically translated into the nodes and edges of the graph, respectively. Graph neural networks generate predicted outlet concentration values by extracting both feature and structural information from the entire graph.

\noindent
\textbf{Graph Matching.} To effectively address the graph matching task, it is imperative to assess the similarity between pairs of graphs by considering their topological structures and node/edge feature information. In numerous scholarly articles, graph matching has been employed to predict the drug-drug interaction (3DProtDTA\cite{voitsitskyi20233dprotdta_directed_app_ddi1}, DGNN-DDI\cite{ma2023dual_directed_app_ddi2}, GMIA\cite{yan2024predicting_directed_app_ddi3}, SISDTA\cite{huang2024structure_directed_app_ddi5}) . In this context, each drug is represented as a molecular graph, wherein nodes symbolize atoms and edges signify the bonds between them. Graph Neural Networks estimate the probability of interactions between two drugs by assessing the similarity between pairs of input graphs.

\section{Future Directions}
\label{sec:future}
Data-centric directed graph learning has emerged as a significant and growing area within the broader field of graph machine learning. In this section, we outline several promising directions for future research in data-centric directed graph learning, aiming to address current challenges and unlock its full potential in diverse application domains.

\subsection{Advancing Industrial Applications of Directed GNNs}
Although Section~\ref{sec:applications} has extensively discussed the industrial applications of directed graph neural networks, we believe that there are still numerous unexplored opportunities for their applications. In the following, we will explore the potential of directed GNNs in industrial contexts from two perspectives: remodeling current applications through the innovative perspective of graph neural networks and extending graph neural networks to additional industrial domains.

\noindent
\textbf{Remodeling Current Applications through the Innovative Perspective of Graph Neural Networks.} Although graph representation learning has developed relatively standardized modeling methodologies for specific problems across various application domains, employing a novel perspective to remodel these domains could potentially lead to enhanced results. Several studies have employed innovative remodeling approaches to address real-world applications such as phishing detection and drug-drug interaction. PEAE-GNN~\cite{huang2024peae_directed_app_pd1}, TSGN~\cite{wang2021tsgn_directed_app_pd4}, EthGAN~\cite{tang2024ethgan_directed_app_pd10} and ATGraph~\cite{kim2023graph_directed_app_pd5} remodel the problem of phishing detection within the Ethereum network as a graph classification task, through extracting a subgraph for each account node to serve as its representation. This approach improves efficiency by minimizing training time and resource consumption, enhancing scalability through the use of ego-graphs for newly identified accounts, and facilitating rapid phishing detection via targeted data acquisition. DGAT-DDI~\cite{feng2022_directed_app_ddi6}, SGRL-DDI~\cite{feng2023social_app_ddi7}, IDDGCN~\cite{gao2024interpretable_directed_app_ddi9} and EmerGNN~\cite{zhang2023emerging_directed_app_ddi8} remodels the drug-drug interaction as an edge level task within a directed graph, rather than framing it as a graph matching problem. It organizes the graph data in the drug-drug interaction networks into a graph network where nodes represent drugs and edges denote interactions between them. This method not only predicts the presence of a drug-drug interaction but also forecasts the directionality of such interaction.

\noindent
\textbf{Extending Graph Neural Networks to Additional Industrial Domains.} Beyond the application areas discussed in Table~\ref{tab:DGNN_app}, directed GNNs are anticipated to hold significant potential for a wider range of industrial applications, such as supply/inventory prediction, wind speed forecasting and impact force identification. More specifically, to achieve accurate supply and inventory forecasting, a supply chain graph can be constructed with factories, distribution centers, and retailers represented as nodes, and supply-demand relationships as edges. This approach formulates the problem as an edge regression task within the framework of graph learning. Moreover, in the domain of wind speed forecasting, a directed graph can be established with turbines as nodes and edges depicting the correlations of wind speeds among different turbines. GNNs can then be utilized to predict the wind speed at each node, leveraging the structured data within this graph-based framework. To simultaneously identify the time and location of the impact force, a directed graph can be constructed wherein sensors function as nodes and the links between sensors serve as edges, with the direction of the edges denoting the temporal sequence. A GNN can then analyze the entire graph to deliver predictions.

\subsection{Enhancing Directed GNNs from the Data-centric Perspective}
As discussed in Section~\ref{sec:revisit}, existing directed GNNs have employed various data enhancement techniques derived from specific graph views to improve performance. However, there remains substantial scope for advancing directed GNNs by further exploring the two core directions of data-centric machine learning: graph data understanding and graph data improvement. Such efforts are essential for fully harnessing the potential of directed GNNs in diverse applications.

\noindent
\textbf{Latent Upgrades from the Perspective of Graph Data Understanding.} As highlighted in Section~\ref{sec:gdue}, most existing directed GNNs predominantly interpret graphs as topological structures or spectral representations. In contrast, relatively few approaches have explored the graph sequence view, which regards graphs as various discretized sequences, leading to the emergence of sequence-based directed graph models only in recent years. This survey emphasizes that significant potential remains for advancing models through the graph sequence lens. For instance, future research could build upon the framework of subgraph-token-based sequences, as demonstrated by DiRW~\cite{su2024dirw}, by designing more effective subgraph token generators and employing more advanced aggregation mechanisms tailored to capture complex dependencies within the sequence representation.

\noindent
\textbf{Latent Upgrades from the Perspective of Graph Data Improvement.} This survey identifies that existing directed GNNs have incorporated several directed graph data enhancement techniques, such as directed multigraphs, directed acyclic graphs, and directed augmented graphs, to improve models' performance, as detailed in Section~\ref{sec:GDI}. However, there remains a notable gap in the research focusing on the enhancement of node features. This survey posits that significant potential still exists for advancing the capabilities of directed GNNs by prioritizing node feature improvement. For instance, further efforts could focus on denoising node features by removing detrimental or irrelevant features, thereby enhancing model's performance. An example of such an approach is demonstrated by MGC~\cite{zhang2021mgc}, which employs feature denoising strategies to achieve superior outcomes after training.

\subsection{Promoting Directed GNNs Beyond Traditional Dimensions}
Future research on directed GNNs can benefit from a unified focus on enhancing integration, scalability/efficiency, robustness, and evaluation benchmarks \& frameworks. These priorities advocate for the development of directed GNN models that are meticulously trained, optimized for resource efficiency, resilient to various challenges, and adaptable to a wide range of application contexts. By addressing these dimensions, future efforts can contribute to the advancement of directed GNNs as versatile and reliable tools for complex graph-based learning tasks. The core aspects are as follow:

\begin{itemize}
    \item \textbf{Integration with Advanced Learning Paradigms:} Integrating advanced learning paradigms, such as self-supervised learning, reinforcement learning, and contrastive learning, offers significant potential to enhance the capabilities of directed GNNs. These approaches can improve the handling of unlabeled data, accommodate dynamic graph structures, and effectively model complex interdependencies. By leveraging these paradigms, directed GNNs can achieve greater performance across diverse graph-based tasks and application domains.

    \item \textbf{Scalability and Efficiency:} Addressing the computational demands of large-scale directed graphs by leveraging techniques such as graph sparsification, efficient sampling, and distributed processing to maintain performance under resource constraints scenarios.

    \item \textbf{Robustness and Interpretability:} Designing directed GNNs that can withstand multiple distribution shifts, adversarial perturbations and noisy data, alongside providing transparent mechanisms for interpreting the directed flow of information and feature relevance within directed graphs.

    \item \textbf{Evaluation Benchmarks \& Frameworks:} Developing comprehensive benchmarks and standardized evaluation metrics tailored to the unique properties of directed graphs is critical for ensuring consistent and reliable comparisons across diverse datasets and downstream applications. Such benchmarks should account for the asymmetry and directionality inherent in directed graphs while providing robust frameworks for assessing model performance across a variety of tasks.
\end{itemize}

\section{Conclusion}
\label{sec:conclusion}
This survey presents a comprehensive review and forward-looking perspective on directed graph machine learning, approached from a data-centric viewpoint. It introduces a novel taxonomy for classifying directed GNNs, encompassing the message-passing framework, the eigenpolynomial framework, and the sequence-based framework. Additionally, the survey revisits directed GNNs through the lens of data-centric methodologies, emphasizing the importance of interpreting and refining graph data to support the development of more expressive and effective GNN models.  
The survey further explores the broad applications of directed GNNs across diverse real-world domains, spanning financial network analysis to brain network analysis, and highlights their significant impact in these areas. Finally, it identifies critical opportunities and challenges in the field of directed graph learning, offering valuable insights to guide future research and development within the graph learning community. Overall, this work underscores the pivotal role of data-centric directed graph machine learning as a foundational area for advancing graph machine learning.

\newpage
\bibliographystyle{ACM-Reference-Format}
\bibliography{reference}

\end{document}